\documentclass{article}

\usepackage{arxiv}
\usepackage{natbib}
\usepackage{hyperref}       
\usepackage{url}            
\usepackage{booktabs}       
\usepackage{amsfonts}       
\usepackage{nicefrac}       
\usepackage{microtype}      
\usepackage{xcolor}         
\usepackage{amsmath}
\usepackage{graphicx}
\usepackage{subcaption}

\usepackage{algorithm}
\usepackage{algpseudocode}
\usepackage{multirow}
\usepackage{float}

\usepackage{adjustbox}

\usepackage{amsmath}
\usepackage{tikz,pgfplots}
\usetikzlibrary{calc, positioning,arrows.meta,shapes,fit}
\pgfplotsset{compat=1.18}

\newtheorem{theorem}{Theorem}[section]
\newtheorem{lemma}[theorem]{Lemma}
\newtheorem{corollary}[theorem]{corollary}
\newtheorem{remark}[theorem]{remark}
\newtheorem{assumption}[theorem]{assumption}

\newtheorem{proof}{Proof}[section]
\title{Fisher Information for Robust Federated Cross-Validation}

\author{ \href{https://orcid.org/0000-0003-0985-9543}{\includegraphics[scale=0.06]{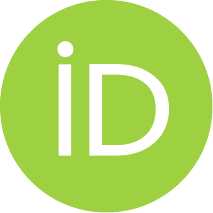}\hspace{1mm}Behraj Khan}\thanks{Both authors contributed equally to this work} \\
	School of Mathematics and Computer Science\\
	Institute of Business Administration Karachi\\
	Pakistan\\
	\texttt{behrajkhan@gmail.com} \\
	\And
     \href{https://orcid.org/0000-0003-0638-9689}{\includegraphics[scale=0.06]{orcid.pdf}\hspace{1mm}Tahir Qasim Syed}\thanks{Both authors contributed equally to this work} \\
	School of Mathematics and Computer Science\\
	Institute of Business Administration Karachi\\
	Pakistan\\
	\texttt{tahirqsyed@gmail.com} \\
}

\renewcommand{\shorttitle}{FIRE}

\hypersetup{
pdftitle={A template for the arxiv style},
}

\begin{document}
\maketitle

\begin{abstract}

When training data are fragmented across batches or federated-learned across different geographic locations, trained models manifest performance degradation. That degradation partly  owes to  covariate shift induced by data having been fragmented across time and space and producing  dissimilar empirical training distributions. Each fragment's distribution is slightly different to a hypothetical unfragmented training distribution of covariates, and to the single validation distribution.  To address this problem, we propose Fisher Information for Robust fEderated validation (\textbf{FIRE}). This  method  accumulates  fragmentation-induced covariate shift divergences from the global training distribution  via an approximate Fisher information. That term, which we prove to be a more computationally-tractable estimate, is then used as a per-fragment loss penalty, enabling scalable distribution alignment.  FIRE outperforms importance weighting benchmarks by $5.1\%$ at maximum and federated learning (FL) benchmarks by up to $5.3\%$ on shifted validation sets.

\end{abstract}

\section{Introduction}

Machine learning models and systems demonstrate strong predictive performance for cross-sectional data as long as the test data distribution aligns with the training distribution. However, these systems often fail to generalize when the test distribution diverges from  training distribution \cite{hendrycks2019benchmarking, taori2020measuring}. These learning-based systems play a critical role in real-world decision-making scenarios. For instance, a pneumonia detection model trained on chest X-rays from specific hospitals may perform poorly when deployed in new geographic regions due to covariate shift (features training distribution \(P_{tr}(x)\) differs from test distribution \(P_{tst}(x)\)) \cite{gardner2023benchmarking}. Similarly, fraud detection systems may struggle to adapt across regions where fraud patterns vary significantly \cite{guan2024federated}. Such failure occurs because standard cross-validation assumes that data is independently and identically distributed \cite{moreno2012study}. The assumption does not hold in real-world as data-sets evolve  increasingly  \textit{fragmented across time, location or devices}, a phenomenon which is referred as \textbf{fragmentation induced covariate shift (FICS)} \cite{khan2025causal}. 

While covariate shift is a well-studied problem in machine learning \cite{sugiyama2007covariate}, however, it is still under-investigated in regimes where it is induced by fragmentation \cite{moreno2012study,sugiyama2007covariate,khan2025causal,khan2025technical}. Importance weighting-based methods such as importance weighted cross validation (IWCV) \cite{sugiyama2007covariate}, density ratio estimation (uLSIF) \cite{kanamori2012statistical}, direct importance estimation \cite{sugiyama2007direct}, dynamic importance weighting (DIW) \cite{fang2020rethinking}, and generalized importance weighting (GIW) \cite{fang2023generalizing} where machine learning models are trained by assigning weights \(w(x)\) to each training example \(w(x) =  \frac{p_{tst}(x)}{ p_{trn}(x)}\) assumes a single source distribution
 and fine-tuned methods uses feature alignment technique \cite{ganin2015unsupervised},  while federated learning approaches ignore local shift during validation process \cite{kairouz2021advances}. 

 To our knowledge, there are no methods which consider covariate shift caused by fragmentation where batches diverge from the validation set and among each other as well.

Non-IID  ( (non-
independent and identically distributed) data in FL is well-studied~\cite{mcmahan2017communication, lu2024federated}. Classical methods (FedAvg, FedProx, SCAFFOLD, FedDyn) focus on stabilizing convergence under heterogeneity, while recent work (e.g., MOON~\cite{li2021model}) aligns local and global representations to mitigate client drift. In contrast, FICS targets robustness to a distinct validation distribution ($P_{val}$), a challenge overlooked by prior approaches.

Domain generalization (DG) methods such as \textsc{Fishr}~\cite{rame2022fishr} also leverage the Fisher Information Matrix (FIM), but with a different goal, enforcing invariance across multiple source domains to generalize to unseen targets. In contrast, FIRE aligns models trained on fragmented batches or clients with a fixed validation distribution ($P_{val}$). Specifically, we penalize parameter sensitivities via Fisher penalty in the direction of $P_{val}$, ensuring validation consistent adaptation. Unlike \textsc{Fishr}, which matches FIMs \emph{across} domains, FIRE compares them \emph{against} validation, making it the first method to explicitly handle fragmentation in both batch/fold processing and federated learning.

Recent FL methods address diverse forms of heterogeneity, like \textsc{LfD}\cite{kim2023learning} regularizes update directions to reduce client drift, \textsc{FedAS}\cite{yang2024fedas} aligns parameters to handle intra/inter-client inconsistency, and \textsc{FedCFA}~\cite{jiang2025fedcfa} mitigates Simpson’s paradox via counterfactual samples. While effective, these approaches optimize within-training consistency and overlook robustness to a fixed validation distribution. FIRE instead introduces a Fisher-based regularizer that explicitly aligns each client or batch/fold with $P_{val}$, addressing fragmentation-induced covariate shift.

In summary, FIRE offers a new perspective on distribution shift by leveraging Fisher information for validation alignment under fragmentation. To our knowledge, it is the first unified framework that addresses fragmented batches and federated clients, enabling scalable mitigation of covariate shift relative to a fixed validation distribution. The working mechanism of FIRE is given in following Figure \ref{fig:combined}.

\color{black}
\begin{figure*}[!ht] 
\centering

\begin{tikzpicture}[>=Stealth, font=\small, scale=0.9, every node/.style={transform shape}]

\node[cylinder, draw, shape border rotate=90, aspect=0.25, minimum height=1.5cm, minimum width=1.5cm, fill=cyan!20] (valbox) at (0.0,-4.5) { $P_{\text{val}}(x)$};
\node[below=2pt of valbox, align=center] {Validation Set};

\node[cylinder, draw, shape border rotate=90, aspect=0.25, minimum height=1.5cm, minimum width=1.5cm, fill=blue!20] (c1) at (3.5,-2.3) { $P_1(x)$};
\node[below=2pt of c1, align=center] {Client 1};
\node[cylinder, draw, shape border rotate=90, aspect=0.25, minimum height=1.5cm, minimum width=1.5cm, fill=green!20] (c2) at (3.5,-4.75) {$P_2(x)$};
\node[below=2pt of c2, align=center] {Client 2};
\node[cylinder, draw, shape border rotate=90, aspect=0.25, minimum height=1.5cm, minimum width=1.5cm, fill=orange!20] (c3) at (3.5,-7.75) { $P_3(x)$};
\node[below=2pt of c3, align=center] {Client 3};
\node[cylinder, draw, shape border rotate=90, aspect=0.25, minimum height=1.5cm, minimum width=1.5cm, fill=red!20] (c4) at (12.5,-9.5) { $P (x)$};
\node[below=2pt of c4, align=center] {Input};

\draw[->, thick, dashed] (valbox.east) -- (c1.west);
\draw[->, thick, dashed] (valbox.east) -- (c2.west);
\draw[->, thick, dashed] (valbox.east) -- (c3.west);

\begin{scope}[xshift=5.8cm, yshift=-3.95cm]
\def\nodesize{0.05cm}
\def\layersep{.75}

\foreach \i in {1,...,2} {
  \node[circle, draw, minimum size=\nodesize] (I1\i) at (0,\i+0.5) {};
}

\foreach \i in {1,...,3} {
  \node[circle, draw, minimum size=\nodesize] (H1\i) at (\layersep,\i) {};
}

\node[circle, draw, minimum size=\nodesize] (O1) at (2*\layersep,2) {};

\foreach \i in {1,...,2} {
  \foreach \j in {1,...,3} {
    \draw[->, thick] (I1\i) -- (H1\j);
  }
}
\foreach \i in {1,...,3} {
  \draw[->, thick] (H1\i) -- (O1);
}
\end{scope}

\begin{scope}[xshift=5.8cm, yshift=-6.5cm]
\def\nodesize{0.05cm}
\def\layersep{.75}

\foreach \i in {1,...,2} {
  \node[circle, draw, minimum size=\nodesize] (I2\i) at (0,\i+0.5) {};
}

\foreach \i in {1,...,3} {
  \node[circle, draw, minimum size=\nodesize] (H2\i) at (\layersep,\i) {};
}

\node[circle, draw, minimum size=\nodesize] (O2) at (2*\layersep,2) {};

\foreach \i in {1,...,2} {
  \foreach \j in {1,...,3} {
    \draw[->, thick] (I2\i) -- (H2\j);
  }
}
\foreach \i in {1,...,3} {
  \draw[->, thick] (H2\i) -- (O2);
}
\end{scope}

\begin{scope}[xshift=5.8cm, yshift=-9.5cm]
\def\nodesize{0.05cm}
\def\layersep{.75}

\foreach \i in {1,...,2} {
  \node[circle, draw, minimum size=\nodesize] (I3\i) at (0,\i+0.5) {};
}

\foreach \i in {1,...,3} {
  \node[circle, draw, minimum size=\nodesize] (H3\i) at (\layersep,\i) {};
}

\node[circle, draw, minimum size=\nodesize] (O3) at (2*\layersep,2) {};

\foreach \i in {1,...,2} {
  \foreach \j in {1,...,3} {
    \draw[->, thick] (I3\i) -- (H3\j);
  }
}
\foreach \i in {1,...,3} {
  \draw[->, thick] (H3\i) -- (O3);
}
\end{scope}

\def\cell{0.15cm}
\foreach \i/\y in {1/-2.0, 2/-4.5, 3/-7.5} {
  \coordinate (G\i) at (8.0,\y-0.45);
  \foreach \a in {0,...,5}{
    \foreach \b in {0,...,5}{
      \pgfmathtruncatemacro{\tone}{20+(\a+2*\b)*8}
      \path[fill=orange!\tone, draw=white]
        ($(G\i)+(\a*\cell,\b*\cell)$) rectangle ++(\cell,\cell);
    }
  }
  \draw[black, thick] ($(G\i)+(0,0)$) rectangle ++(6*\cell,6*\cell);
  \node at ($(G\i)+(0.45,1.25)$) {$I_{k}(\theta)$};
}

\foreach \i in {1,2,3} {
  \draw[->, thick] (O\i) -- ($(G\i)+(0,0.45)$);
}

\draw[->, thick, shorten >=5pt, shorten <=5pt] (4.50,-1.95) -- (6.0,-1.95);
\draw[->, thick, shorten >=5pt, shorten <=5pt] (4.50,-4.5) -- (6.0,-4.5);
\draw[->, thick, shorten >=5pt, shorten <=5pt] (4.50,-7.45) -- (6.0,-7.45);

\node[draw, thick, rounded corners, minimum width=3cm, minimum height=4.5cm,
      inner sep=8pt, label=below:{Global Model}] (globalbox) at (11.75,-4.5) {};

\def\cell{0.18cm} 
\coordinate (FIMglobal) at (11.2,-3.47); 

\foreach \a in {0,...,5}{
  \foreach \b in {0,...,5}{
    \pgfmathtruncatemacro{\tone}{20+(\a+2*\b)*6}
    \path[fill=blue!\tone, draw=white]
      ($(FIMglobal)+(\a*\cell,\b*\cell)$) rectangle ++(\cell,\cell);
  }
}
\draw[black, thick] ($(FIMglobal)+(0,0)$) rectangle ++(6*\cell,6*\cell);

\node[text=blue, above=38pt] at ($(FIMglobal)+(3*\cell,-0.2)$) { \textbf{$I_G(\theta)= \sum_{k=1}^{N} I_k(\theta)$}};

\begin{scope}[shift={(11.0cm,-7.5cm)}] 
  \def\nodesize{0.05cm}
  \def\layersep{.75}

  \foreach \i in {1,...,2} {
    \node[circle, draw, minimum size=\nodesize] (I\i) at (0,\i+0.5) {};
  }

  \foreach \i in {1,...,3} {
    \node[circle, draw, minimum size=\nodesize] (H\i) at (\layersep,\i) {};
  }

  \node[circle, draw, minimum size=\nodesize] (O) at (2*\layersep,2) {};

  \foreach \i in {1,...,2} {
    \foreach \j in {1,...,3} {
      \draw[->, thick] (I\i) -- (H\j);
    }
  }

  \foreach \i in {1,...,3} {
    \draw[->, thick] (H\i) -- (O);
  }

  \draw[->, thick] (O) -- ++(2.50,0) node[midway, above] {$\mathcal{L}_{\text{FIRE}}(\theta)$};
\end{scope}

\foreach \i in {1,2,3} {
  \draw[->, thick] ($(G\i)+(6*\cell,0.45)$) -- (globalbox);
}

\node at (11.5,-0.35) { $\mathcal{L}_{\text{FIRE}}(\theta) = \sum_{k=1}^{K} \frac{n_k}{N} \mathcal{L}_k(\theta) + \lambda \cdot (I(\theta))$};

\draw[->, thick, blue] ($(FIMglobal)+(3*\cell,0)$) -- ++(0.0,-0.9);
\draw[->, thick, dashed] (c4.north) -- (globalbox.south);

\draw[xshift = 4.25cm, yshift = -5.25cm,<-,  very thick, dashed, blue!80, scale = 1.25] (0,0) -- (4.75,0);
\draw[<-, xshift = 3.5cm, yshift = -1.75cm, blue, very thick, dashed] 
    (0,0) -- (0,1) -- (8,1) -- (8,0);
\draw[<-, xshift=3.5cm, yshift=-7.95cm, blue, very thick, dashed] 
    (0,0) -- (0,-1) -- (8,-1) -- (8,0);
\draw[xshift=11.5cm, yshift=-6.95cm,blue, very thick, dashed] (0,0) -- (0,-1);

\end{tikzpicture}

    \caption{    \textbf{FIRE} working mechanism in FL setting. The server broadcasts the global model $\theta$ and global FIM $I_G(\theta)$ to all clients. Each client $k$ computes its local FIM $I_k(\theta)$ using the shared validation set $P_{\text{val}}(x)$. Clients perform a local update regularized by $I_G(\theta)$ and send their local FIMs back to the server. The server then aggregates the client FIMs (e.g., $I_G(\theta) = \sum_{k=1}^N \frac{n_k}{N} I_k(\theta)$) to update the global FIM for the next round. This unified approach ensures model alignment with the target validation distribution in both settings.}
    \label{fig:combined}
\end{figure*}
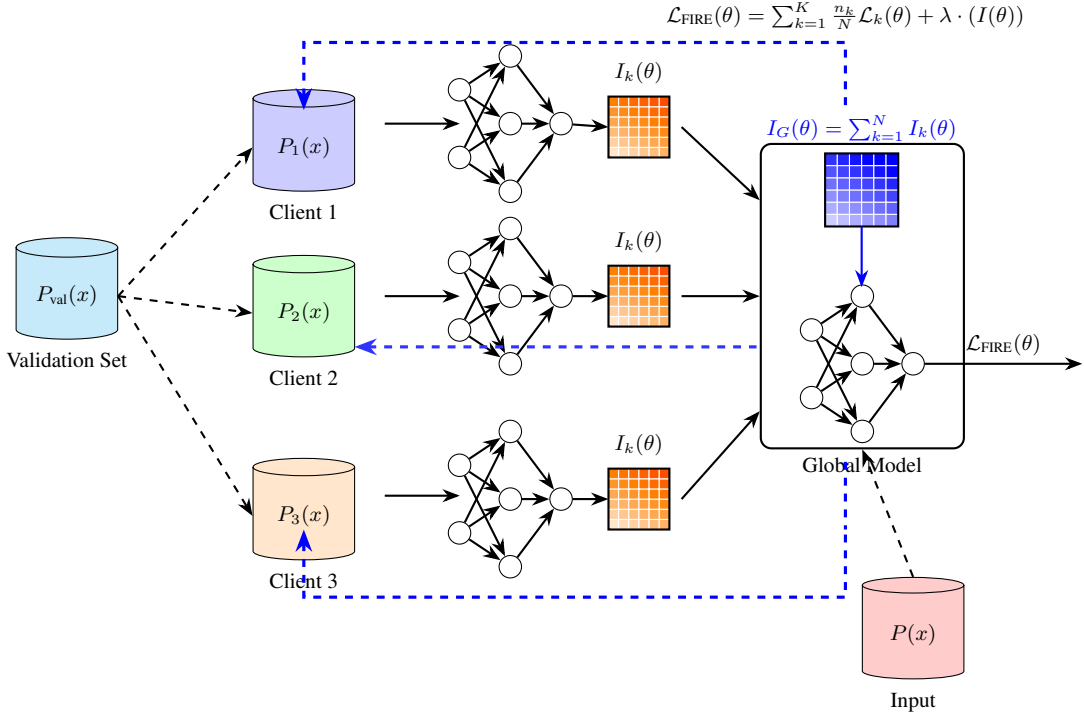
\section*{Contributions}
\begin{enumerate}
    \item We formally define fragmentation-induced covariate shift (FICS) in federated and batch/fold setting, showing  the performance of existing methods is compromised by a non-iid  cross-validation split.
    \item We propose  {FIRE}, the first method to remediate FICS via Fisher information to tractably estimate a function of network parameters. Our method allows memory cost to be linear in the size of a dataset fragment.
    \item We extend FIRE to federated learning with minimum communication overhead, outperforming FL baselines.
\end{enumerate}

\section{Method development}

\subsection{Notational setup}
\label{sec:nsfl}
In FL context, lets assume that there are \(K\) number of clients having their own datasets like: \(\mathcal{D}_{k} = \{ (x_{i}^{k}, y_{i}^{k}) \}_{i=1}^{n_k},\) where \(x_{i}^{k}\)  are the covariates, \( y_{i}^{k}\)  are the labels, while \(n_k\)  are the number of samples for client  \(k\). The data distribution \( P_k(x, y) \) may differ across clients, leading to covariate shift (i.e., \( P_k(x) \neq P_{k'}(x) \) for clients \( k \) and \( k' \)). 

\subsection{Unified Framework for fragmented and federated learning}
\label{sec:unifrmwrk}
Our method, FIRE, handles both traditional fragmented data settings (batches/folds) and federated learning (FL) scenarios under a unified framework. While these two settings differ in data partitioning, they share the core challenge of covariate shift between training and validation distributions. We formalize this connection below:
\begin{itemize}
    \item In fragmented setting, data split into \(k\) batches \(\{B_i\}_{i=1}^m\) with distributions $P_i(x) \neq P_{val}(x)$. Our goal is to minimize validation loss \(\mathcal{L}_{\theta} = \mathbb{E}_{(x,y)\sim P_{val}}[\ell(y, f_\theta(x))]\) by aligning \(P_{i}\) with \(P_{val}\) via FIM.
    \item In FL settings the data is partitioned across \(K\) clients \( \{ D_k \}_{k=1}^K \) with 
\( P_k(x) \neq P_{\text{val}}(x) \). Goal is to minimize the global loss \(\mathcal{L}_{(\theta)} = \sum_{k=1}^{K} \frac{n_k}{N} \mathcal{L}_{k}(\theta)
\)
while ensuring client models generalize to \( P_{\text{val}} \).
\end{itemize}
The common idea in both settings is estimating and mitigating covariate shift.  Fragmented settings computes FIM \(I_{i}(\theta)\) for batches/folds while FL computes it for clients \(I_{k}(\theta)\) to measure \(\mathcal{D}_{KL}(P_{i}||P_{val})\). During the remediation phase, both settings penalize the loss with \(\lambda.I(\theta)\) where \(I(\theta)\) is (batchwise or clientwise) aggregated FIM.  The FIRE algorithm \ref{alg:fire_fixed} applies identically in both cases. For batches, \(I_{G}(\theta)\) accumulates shift across sequential batches while in FL scenario, \(I_{G}(\theta)\) is the weighted average of client FIMs. We assume access to a small public validation set \(V\)
that is representative of the target distribution. For simplicity of exposition, Figure \ref{fig:combined} shows this set being shared with clients. In practice, if sharing raw examples is undesirable, the server can compute the validation FIM once (or periodically) and broadcast a compressed approximation (e.g., diagonal) to clients, thereby preserving privacy.

\begin{algorithm}[t]
\caption{FIRE: Batchwise Fisher accumulation for covariate-shift remediation }
\label{alg:fire_fixed}
\begin{algorithmic}[1]
\Require Batches $\{B_i\}_{i=1}^m$, validation set $V$, learning rate $\eta$, penalty $\lambda$, momentum $\alpha$, mixing weight $\mu \in [0,1]$
\Ensure Robust model parameters $\theta$
\State Initialize $\theta_0$, $I_G \leftarrow \mathbf{0}$
\State \textbf{Precompute (or periodically update) validation FIM:}
\State $I_V(\theta) \leftarrow \mathbb{E}_{(x,y)\sim V}\left[\nabla_\theta \log p(y|x;\theta)\,\nabla_\theta \log p(y|x;\theta)^\top\right]$
\For{each batch $B_i \in \{B_1, \dots, B_m\}$}
    \State Compute batch FIM (mini-batch estimate):
    \State $I_{B_i}(\theta) \leftarrow \mathbb{E}_{(x,y)\sim B_i}\left[\nabla_\theta \log p(y|x;\theta)\,\nabla_\theta \log p(y|x;\theta)^\top\right]$
    \State Form combined per-batch FIM (mix validation and batch):
    \State $I_i(\theta) \leftarrow \mu\, I_{B_i}(\theta) + (1-\mu)\, I_V(\theta)$
    \State Update global FIM with momentum:
    \State $I_G(\theta) \leftarrow \alpha I_G(\theta) + (1-\alpha) I_i(\theta)$
    \State Perform SGD update with FIM-based regularization:
    \State $\theta \leftarrow \theta - \eta \Big( \nabla_\theta \mathcal{L}(B_i) + \lambda \, I_G(\theta) \, \nabla_\theta \mathcal{L}(B_i) \Big)$
\EndFor
\State \Return $\theta$
\end{algorithmic}
\end{algorithm}

\subsection{Federated learning specifics} Now we detail our method FIRE practical considerations communication efficiency, scalability, and comparisons to standard FL baselines ensuring feasibility in real-world deployments.

\textbf{Communication efficiency.} FIRE transmits client FIM to the server once per global round. For a model with $d$ parameters, each client sends $\mathcal{O}(d^2)$ FIM entries (symmetric, so approximately $d^2/2$ values).
 Following \cite{rothchild2020fetchsgd}, we use a rank-$k$ approximation ($k \ll d$) to reduce the overhead to $\mathcal{O}(kd)$. FIRE adds minimal overhead compared to gradient transmission ($\mathcal{O}(d)$ per client), as FIMs are aggregated infrequently (every 5 rounds).

\textbf{Scalability.} The FIRE framework per-client computation scales as:
\(
    \text{FIM Cost} = O(b \cdot d^2),
\)
where b is the batch size. For large d, we approximate \(I_{k}(\theta)\) as diagnol, reducing cost to \(O(b.d)\) \cite{kingma2020method}.

\subsection{Problem formulation}
Let $\mathcal{D} = \{B_i\}_{i=1}^m$ be a fragmented dataset into \(m\) number of batches with batch distribution $P_i(x) \neq P_{val}(x)$, \(P_i(x)\) is an arbitrary batch distribution while \(P_{val}(x)\) is validation set distribution. Our goal is to minimize loss \(\mathcal{L}_{\theta}\):
\begin{equation}
\mathcal{L}_{\theta} = \mathbb{E}_{(x,y)\sim P_{val}}[\ell(y, f_\theta(x))].
\end{equation}
A principled metric over the probability space distribution is required for measuring the amount of shift between fragmented batches and validation set \cite{khan2025causal}. The KL divergence \cite{kullback1951information} is a natural choice for batch comparison and its close connection to cross entropy loss commonly used in neural networks. In practice KL divergence is often used as mean-field approximation, where the posterior \(q(\theta)\) assumed to be Gaussian and parametrized by covariance of networks weights. However, access to the Hessian of the loss with respect to model \(f(\theta)\) is required for computing this term, which is infeasible in high-dimensional settings. To circumvent this, \cite{pascanu2013revisiting} proposed approximation of Hessian  \( \mathbb{E}\left[- \frac{\partial^2 \log p(X\mid\theta)}{\partial\theta\partial\theta^T}\right]\) using FIM (Fisher information matrix) \(I(\theta)\), which is more tractable alternative that also captures the second order information and can be estimated using the expected values and variance of the gradients \cite{nishiyama2019new}.
\subsection{Fisher Information Approximation}
Fisher information for each batch \(B_i\) and validation set \(V\) is computed by:
\begin{equation}
    I_i(\theta) = \mathbb{E}_{x\sim V} \left[ \nabla_\theta \log p(y|x;\theta) \nabla_\theta \log p(y|x;\theta)^T \right].
\end{equation}
where \(\mathbb{E}_{x\sim V}\) is expectation over validation set and \(\nabla_\theta \log p(y|x;\theta) \) is log-likelihood of model \(f(\theta)\) predictions with respect to parameters \(\theta\). \(I_i(\theta)\) approximates the curvature of \( D_{\mathrm{KL}}(P_i \,\|\, P_{\mathrm{val}}) \).


\begin{assumption}[Model regularity and bounds]
\label{ass:regularity}
The conditional model $p(y\mid x;\theta)$ satisfies the following for all $x,y,\theta,\theta'\in\mathbb R^d$:
\begin{enumerate}
  \item[(R1)] (Lipschitz Hessian) $\|\nabla^2_\theta \log p(y\mid x;\theta') - \nabla^2_\theta \log p(y\mid x;\theta)\|_{\mathrm{op}}
  \le \beta \|\theta'-\theta\|_2$.
  \item[(R2)] (Bounded score) $\|\nabla_\theta \log p(y\mid x;\theta)\|_2 \le G$ almost surely; hence the local Fisher satisfies $\|F_x(\theta)\|_{\mathrm{op}}\le G^2=:M$.
  \item[(R3)] (Regularity) For every $x,\theta$ we have $\mathbb E_{y\sim p(\cdot\mid x;\theta)}[\nabla_\theta\log p(y\mid x;\theta)] = 0$.
\end{enumerate}
\end{assumption}

\begin{assumption}[Data proximity]
\label{ass:data}
Let $P_i(x)$ and $P_{\mathrm{val}}(x)$ be two marginal distributions. Define the Radon--Nikodym derivative $r(x)=\frac{dP_i}{dP_{\mathrm{val}}}(x)$ with assumption
\(
|r(x)-1|\le \gamma < 1 \quad\text{for all }x.
\)
\end{assumption}

\begin{lemma}[Marginal KL bound]
\label{lemma:marginalKL}
Under Assumption~\ref{ass:data} we have
\(
D_{\mathrm{KL}}(P_i(x)\|P_{\mathrm{val}}(x))
= \mathbb E_{x\sim P_{\mathrm{val}}}\big[ r(x)\log r(x)\big]
\le C_1 \gamma^2 + C_1' \gamma^3,
\)
where one can take for instance
\(
C_1=\frac{1}{2(1-\gamma)},\qquad C_1'=\frac{1}{3(1-\gamma)^2}.
\) see proof detail in appendix \ref{lemma:appmarginalKL}
\end{lemma}

\begin{lemma}[Local conditional KL quadratic expansion]
\label{lemma:localKL}
Fix $x$. Under Assumption~\ref{ass:regularity} and for two parameter vectors $\theta_i,\theta_{\mathrm{val}}$ with $\|\theta_i-\theta_{\mathrm{val}}\|_2\le \delta$, the conditional KL admits the expansion
\(
D_{\mathrm{KL}}\big(p(\cdot\mid x;\theta_i)\,\big\|\,p(\cdot\mid x;\theta_{\mathrm{val}})\big)
= \tfrac12(\theta_i-\theta_{\mathrm{val}})^\top F_x(\theta_i)(\theta_i-\theta_{\mathrm{val}})
+ R_x,
\)
with the remainder bounded by
\(
|R_x|\le \frac{\beta}{6}\, \delta^3 \, G,
\)
so in particular $|R_x|\le \tfrac{\beta G}{6}\delta^3$ see proof detail in appendix \ref{lemma:applocalKL}.
\end{lemma}

\begin{theorem}[KL divergence bound via Fisher information]
\label{thm:KL-via-Fisher}
Suppose Assumptions~\ref{ass:regularity} and \ref{ass:data} hold. Let $\theta_i,\theta_{\mathrm{val}}$ satisfy $\|\theta_i-\theta_{\mathrm{val}}\|_2\le \delta$. Then
\begin{equation}
\label{eq:mainbound}
\begin{aligned}
D_{\mathrm{KL}}(P_i\|P_{\mathrm{val}})
&\le \tfrac12 (\theta_i-\theta_{\mathrm{val}})^\top F_{\mathrm{val}}(\theta_{\mathrm{val}})(\theta_i-\theta_{\mathrm{val}}) \\
&\quad + C_1\gamma^2 + C_1'\gamma^3
+ C_2\,\gamma\,\delta^2
+ C_3\,\beta G\,\delta^3,
\end{aligned}
\end{equation}
where $F_{\mathrm{val}}(\theta)=\mathbb E_{x\sim P_{\mathrm{val}}}[F_x(\theta)]$, and one may take
\(
C_1=\frac{1}{2(1-\gamma)},\quad C_1'=\frac{1}{3(1-\gamma)^2},\quad C_2=\frac{M}{2},\quad C_3=\frac{1}{6}.
\)
\end{theorem} see proof detail in appendix \ref{thm:appKL-via-Fisher}.

\color{black}

\subsection{Connection to Federated Learning}
\label{sec:cfl}
Fragmentation induced covariate shift occurs when sequence of batches (i.e clients in federated leaning) affects the covariates distributions. In federated learning (FL) the data arrives similarly in non-iid manner across clients where the clients are often at different geographic locations. The data distributions differs due to temporal, geographic or user-specific factors which leads to covariate shift  \cite{mcmahan2017communication,du2022rethinking,ramezani2023federated}. In FL setting, the model trained for one client data may not be able to generalize to other client data. Prior FL methods such as FedProx and  SCAFFOLD \cite{karimireddy2020scaffold} try to mitigate client shift via regularization or variance reduction, they ignore validation-time shift—the misalignment between a client’s local data and the global validation set. In our method we address this problem by Fisher information penalty which can be applied in  FL setting as well for model generalization. As in batchwise setting, the knowledge about data density of training batch is accumulated to penalize the loss in subsequence batches. In FL as the data is distributed across clients, the same approach can be applied here by accumulating knowledge about data density of each client can be used by the global model for correction of covariate shift across clients. The goal of FIRE in FL setting where data is distributed across clients, is to adapt to covariate shift while training the global model. 

\begin{equation}
    \mathcal{L}(\theta) = \sum_{k=1}^{K} \frac{n_k}{N} \mathcal{L}_k(\theta)
\end{equation}

 where \(\mathcal{L}_k(\theta)\) is loss of local client and  \(N =\sum_{k=1}^{K}{n_k} \) is number of samples across all clients.

 In FL, the global model leads to poor generalization due to covariate shift arises by the change in features distributions \(p(x)\) across clients. FIRE corrects this covariate shift by penalizing the loss function with Fisher information penalty like batch/fold setting. In FL setting, FIRE computes the FIM \(I_K(\theta)\) for each client \(k\) distribution i.e
\(
I_k(\theta) = \mathbb{E}_{(x,y) \sim P_k(x,y)} \left[ -  \frac{\partial^2 \log p(y \mid x; \theta)}{\partial \theta \partial \theta^2} \right]
\).
 Th term \(I_k(\theta)\)  provides information about local client \(k\) data distribution and captures the curvature of local loss function \(\mathcal{L}_k(\theta)\). This term is then integrated into local client loss as penalty term such as:
 \begin{equation}
 \mathcal{L}_k(\theta) = \mathcal{L}_k(\theta) + \lambda \cdot (I_k(\theta))
\end{equation}
 After computing FIM for all clients, global model then find weight average of these local FIMs as:
 \begin{equation}
 I(\theta) = \sum_{k=1}^{K} \frac{n_k}{N} {I}_k(\theta)\end{equation}
 The global FIM contain curvature of all local clients loss.  In FL, this term is integrated into global model loss function as penalty term to correct for covariate shift i.e:
 \begin{equation}
 \mathcal{L}_{\text{FIRE}}(\theta) = \sum_{k=1}^{K} \frac{n_k}{N} \mathcal{L}_k(\theta) + \lambda \cdot (I(\theta)) \end{equation}

 where \(\lambda\) is hyper-parameter for controlling strength of penalty. In optimization phase, the clients computes its local FIM \(I_k(\theta)\) and local gradient \(\nabla \mathcal{L}_k(\theta)\) first then the global model aggregates these FIMs and local gradients for the global update. 
 \begin{equation}
 \theta \leftarrow \theta - \eta \left( \sum_{k=1}^{K} \frac{n_k}{N} \nabla L_k(\theta) + \lambda \cdot \nabla (I(\theta)) \right)
\end{equation}
where \(\eta\) is global model learning hyper-parameter.
The introduced Fisher information acts as regularizer which penalizes the model parameters which leads to covariate shift. It helps the global model in better generalization across the clients with different data distributions. The integration of this penalty into global model make it robust to covariate shift, which is common challenge in FL settings due to non-iid nature of client data. The results show in Table \ref{tab:fl_results} provide empirical evidence to the effectiveness of this method.
\section{Related work}
\textbf{Covariate shift.} In supervised machine learning the model expects that sample test distribution follows same training distribution \cite{vapnik1998statistical, scholkopf2002learning, duda2006pattern}. However, this assumption does not hold in real-world due to non-stationary environment or samples bias selection \cite{quinonero2022dataset,sugiyama2012machine}. The term \textit{Covariate shift} was coined by Shimodaira \cite{shimodaira2000improving} where covariates (features) training distribution differs from test distribution. Covariate shift is common in many real-world applications, such as emotion recognition \cite{jirayucharoensak2014eeg} speaker identification \cite{yamada2010semi} and brain-computer interface \cite{li2010application}.

\textbf{Importance weighting.} Importance weighting (IW) is the most common approach used for adaptation under distribution shift. IW estimates density ratio between training and test distribution and uses it for reweighting the training loss during optimization \cite{shimodaira2000improving, sugiyama2007direct}. Kernel mean matching (KMM) \cite{gretton2009covariate} minimize maximum mean discrepancy (MMD) in kernel Hilbert space to align training and test distribution.  \cite{sugiyama2007direct} uses direct density ratio estimation KLIEP for covariate shift adaptation.  Recent methods like dynamic importance weighting (DIW) \cite{fang2020rethinking} address distribution shift by updating the importance weights during stochastic optimization while avoiding the offline density ratio estimation. Generalize importance weighting (GIW) \cite{fang2023generalizing} adapt to distribution shift by dynamically estimating the importance weights for training examples using gradient-based optimization without explicit density ratio estimation. 

\textbf{Domain Adaptation.} In domain adaptation, distribution shift is addressed by aligning the source and target domains with assumption that target data is accessible during training. Adversarial methods such as DANN \cite{ganin2016domain} learn domain-invariant features using gradient reversal, while discrepancy-based methods minimize divergence metrics such as MMD \cite{tzeng2014deep} or CORAL \cite{sun2016deep}. Recent work, such as \cite{zhao2018learning}, extends domain adaptation to multi-source settings or partial adaptation \cite{cao2022big}. However, these methods fail when target data is inaccessible (e.g. fragmented batches) or when distribution shift occurs across clients in federated learning (FL) settings.

\textbf{Federated learning under covariate shift.} 
Non-IID client data induces covariate shift in FL, challenging global model training. FedAvg~\cite{mcmahan2017communication} averages local updates but struggles with client-specific shifts; extensions such as FedBN~\cite{li2021fedbn}, FedProx~\cite{li2020federated}, clustering~\cite{zhang2020personalized}, and meta-learning~\cite{jiang2019improving} improve stability under heterogeneity. Contrastive methods like MOON~\cite{li2021model} reduce client drift, while recent advances (LfD~\cite{kim2023learning}, FedAS~\cite{yang2024fedas}, FedCFA~\cite{jiang2025fedcfa}) address drift, inconsistency, and aggregation bias. Yet, these approaches focus on training- or aggregation-time robustness and overlook distribution shift at \emph{validation} time. Our method closes this gap by introducing Fisher-driven alignment with a fixed validation distribution.

\color{black}

\section{Experiments}

We evaluated the effectiveness of our method \textbf{FIRE} in fragmented (batches) and in federated settings (clients) on standard benchmarks. 

\textbf{Datasets.} FIRE performance evaluated on 39 total datasets. The datasets includes: {F-MNIST} \cite{xiao2017fashion}, {K-MNIST} \cite{clanuwat2018deep} and  {MNIST-C} \cite{mu2019mnist}  which is also used by Sugiyama et. al, as benchmark for covariate shift adaptation . For inducing shift in these three datasets, we follow the same procedure given in  One-step approach \cite{zhang2020one}\footnote{Each training image $I_i$ is rotated by angle $\theta_i$, with $\theta_i/180^\circ$ drawn from distribution $\text{Beta}(a, b)$. For test images $J_i$, the rotation angle $\phi_i$ is drawn from $\text{Beta}(b, a)$  $= (2, 4)$, $(2, 5)$, and $(2, 6)$.} Covariate shift is induced on five tabular datasets—Australian, Breast Cancer, Diabetes, Heart, and Sonar from the KEEL repository \cite{alcala2011keel}, following the procedure of Sugiyama et al. \cite{zhang2020one}, originally adapted from Cortes et al. \cite{cortes2008sample}. All other datasets, including image datasets such as MNIST \cite{lecun2010mnist}, EMNIST \cite{cohen_afshar_tapson_schaik_2017}, FEMNIST, EMNIST-D QMNIST \cite{qmnist-2019}, Kannada-MNIST \cite{prabhu2019kannada}, CIFAR-10  and CIFAR-100 \cite{coates2011analysis}, SVHN \cite{Netzer2011}, P-MNIST \cite{mu2019mnist}, and corruption variants CIFAR-10-C and CIFAR-100-C \cite{hendrycks2019robustness}, as well as 27 additional binary classification datasets from KEEL \cite{alcala2011keel}, are used under their standard published settings for evaluating FIRE performance.

\textbf{Model architecture.} In fragmented (batches/folds) setting, we use a five-layer convolutional neural network (CNN) with softmax cross-entropy loss for all image-based benchmarks. The architecture consist of two convolutional layers with pooling followed by three fully connected layers. Hyperparameters (optimizer = Adam, activation = softmax, and training epochs = 100) are fixed across image datasets. For tabular datasets, we employ a consistent multi-layer perceptron (MLP) architecture with a single hidden layer of 4 neurons. The tabular setting uses hyperparameters (activation = ReLU, optimizer = Adam, and epochs = 1500). All reported accuracies are averaged over 100 independent runs.

In federated learning setting we used a three-layer fully connected neural network with ReLU activations, mapping 784-dimensional inputs to 10 output classes via hidden layers of size 512 and 256. This architecture is used consistently across all clients. The model is trained with  FIM penalty to mitigate  covariate shift in FL setting.

The penalty coefficient $\lambda$ is held constant across all datasets, calibrated using a batch/fold configuration as shown in Figure~\ref{fig:main}.  

All baselines are implemented in TensorFlow 2.11, and code is available at the anonymous hyper-link\footnote{\href{https://anonymous.4open.science/r/FIRE-7D3B/FIRE}{\textcolor{blue}{FIRE}}}. We reproduce baseline results as reported in the original publications More implementation details can be found in appendix \ref{impdetal}.

\textbf{Evaluation metrics.} For performance evaluation of our method FIRE we used accuracy as a metric. The accuracy metric remains consistent across all our experiments like in fragmented setting (image/tabular), and in FL setting too. 

\textbf{Experimental design.} We evaluate FIRE on fragmented data (batches/folds) and in FL settings (clients) using five sets of experiments, with standard cross-validation (st-CV) as the baseline. The experiments are:
\begin{itemize}

 \item \textbf{Covariate shift in batch settings:} We evaluate st-CV on both the integral dataset and its fragmented (batched) versions to examine the impact of fragmentation-induced covariate shift. The experiments are conducted on 13 image-based and. Results are reported in Table~\ref{tab: bst-CV}.
\item \textbf{FIRE shift mitigation in batch settings:} We apply FIRE with FIM-based penalty on both integral and fragmented batches to evaluate its effectiveness in mitigating covariate shift. The results are presented in Table \ref{tab: ccca}.

\item \textbf{Covariate shift in fold settings:} To assess the impact of fragmentation-induced covariate shift in fold settings, we conduct experiments on 26 tabular datasets. Results are presented in Appendix Table~\ref{tab: c3keel}.

\item \textbf{FIRE shift mitigation in fold settings:} We assess FIRE's efficacy in mitigating covariate shift in fold settings using tabular data Results can found in Table \ref{tab: c3keel}.

\item \textbf{Comparison of FIRE with FL state-of-the-art:} We evaluate FIRE's robustness against state-of-the-art (SOTA) methods like FedAvg \cite{mcmahan2017communication}, SCAFFOLD \cite{karimireddy2020scaffold}, MOON \cite{li2021model}, LfD \cite{kim2023learning}, FedAS \cite{yang2024fedas}, and FedCFA \cite{jiang2025fedcfa}. Results are presented in Table~\ref{tab:fl_results}.

\end{itemize}

\section{Results and Discussion}

\subsection{Covariate shift and batches/folds settings} Table \ref{tab: bst-CV} shows that dataset fragmentation into batches consistently degrades both average and batchwise accuracy compared to the st-CV baseline, due to the covariate shift it induces. On image-based datasets the fragmentation leads to over 36\%  and 60\% drop in average accuracy across 2, 10, and 20 batches,  indicating an effect from induced shift.

It can be noticed from table that effect of fragmentation frequency also effects accuracy. Figure in appendix \ref{fig:batchfrequency} and Table \ref{tab: bst-CV} show that accuracy loss increases with fragmentation frequency such as 52.1\% for 20 batches, 43.7\% for 10, and 36.3\% for 2. Fewer batches offer greater data support, partially mitigating the shift.

st-CV results under varying fold settings are reported in Table~\ref{tab: st-cvfold}. In fold settings the baseline remains same (i.e st-CV) to test whether data fragmentation induces distribution shift. As shown in Table ~\ref{tab: st-cvfold}, the accuracy consistently degrades with increasing folds, indicating shift. We report mean accuracy for each fold setting; $\mu_3$, $\mu_4$, and $\mu_5$ denote averages for (2, 5, and 10) folds, respectively.

\subsection{FIRE shift mitigation in batch/folds settings} 

\textbf{FIRE mitigation in batch settings.} FIRE, effectively mitigates FICS in no-shift settings. As shown in Table~\ref{tab: ccca}, column $\Delta_3$, it improves average accuracy by over 10\% across batch fragmentation levels (20, 10, 2), consistently across datasets. This gain stems from FIRE ability to extract and retain batch-sequence information while regularizing the model. Cross-batch comparisons show significant improvements in remediated shifts when aligned at the same sequence index. For example, under 20-way fragmentation on F-MNIST, batch $B_n$ improves from 70.6\% (with shift, Table~\ref{tab: bst-CV}) to 81.9\% (remediated, Table~\ref{tab: ccca}), a 16\% increase. These results highlight FIRE robustness across batch positions and fragmentation settings.

\textbf{FIRE mitigation in fold settings.} Table~\ref{tab: st-cvfold} presents st-CV results under $k$-fold settings, where st-CV serves as the baseline to test our hypothesis that data the fragmentation induces distribution shift. As shown in Tables~\ref{tab: c3keel} and \ref{tab: st-cvfold}, accuracy degrades with increasing fold count, confirming that finer fragmentation amplifies shift. We report mean accuracies for each setting: $\mu_3$, $\mu_4$, and $\mu_5$ correspond to (2, 5, and 10) folds, respectively.

Tables~\ref{tab: c3keel} and \ref{tab: st-cvfold} present $k = 2, 5, 10$ fold results for tabular datasets. In Table~\ref{tab: c3keel}, accuracy gaps $\Delta_5 = \mu_3 - \mu_6$, $\Delta_6 = \mu_4 - \mu_7$, $\Delta_7 = \mu_5 - \mu_8$ highlight the impact of induced shift. Our method improves accuracy by up to 28.3\% across all settings.

\begin{table*}[!ht]
\scriptsize
\centering
\caption{FIRE benchmarking with SOTA on image datasets ($\boldsymbol\Delta_1$ \% = FIRE - SOTA)}
\label{tab:sota}
\begin{tabular}{ccccccl} 
\hline
\textbf{\textbf{\textbf{\textbf{Dataset}}}} & \begin{tabular}[c]{@{}c@{}}\textbf{\textbf{Shift Level }}\\\textbf{\textbf{(a, b)}}\end{tabular} & \textbf{ERM} & \textbf{EIWERM} & \textbf{One-step} & \textbf{FIRE} & $\boldsymbol\Delta_1$\%  \\ 
\hline
\multirow{3}{*}{F-MNIST}                    & (2, 4)                                                                                           & 64.6 ± 0.17   & 71.3 ± 0.06      & 74.5 ± 0.08        & 78.3 ± 0.14     & $\uparrow$ 5.10\%      \\                                            & (2, 5)                                                                                           & 54.5 ± 0.54   & 57.9 ± 0.29      & 55.6 ± 0.20        & 57.2 ± 0.31     & $\uparrow$ 2.87\%      \\                                            & (2, 6)                                                                                           & 36.3 ± 0.34   & 42.5 ± 0.55      & 44.8 ± 0.25        & 45.4 ± 0.20     & $\uparrow$ 1.33\%      \\ 
\hline
\multirow{3}{*}{K-MNIST}                    & (2, 4)                                                                                           & 67.1 ± 0.18   & 69.7 ± 0.24      & 68.8 ± 0.12        & 72.2 ± 0.44     & $\uparrow$ 4.94\%      \\                                            & (2, 5)                                                                                           & 55.0 ± 0.26   & 52.2 ± 0.19      & 59.5 ± 0.16        & 61.6 ± 0.09     & $\uparrow$ 3.52\%      \\                                            & (2, 6)                                                                                           & 39.2 ± 0.30   & 38.4 ± 0.93      & 43.1 ± 0.55        & 43.8 ± 0.35     & $\uparrow$ 1.62\%      \\ 
\hline
\multirow{3}{*}{MNIST-C}                    & (2, 4)                                                                                           & 63.6 ± 0.91   & 80.5 ± 0.08      & 85.2 ± 0.17        & 86.3 ± 0.28     & $\uparrow$ 1.29\%      \\                                            & (2, 5)                                                                                           & 43.8 ± 0.18   & 60.4 ± 0.47      & 78.4 ± 0.32        & 79.3 ± 0.67     & $\uparrow$ 1.14\%      \\                                            & (2, 6)                                                                                           & 33.3 ± 0.49   & 53.8 ± 0.13      & 64.2 ± 0.67        & 64.9 ± 0.37     & $\uparrow$ 1.09\%      \\
\hline
\end{tabular}
\end{table*}

\begin{table*}[!ht]
\scriptsize
\centering
\caption{Benchmarking with SOTA on tabular datasets. Mean accuracy with \textit{Wilcoxon signed-rank test} \cite{wilcoxon1992individual} at significance level 5\% across various datasets with induced covariate shift ($\Delta_2$ \% = FIRE - SOTA)}
\label{tab:performance}
\begin{tabular}{lcccccc} 
\hline
\textbf{Dataset}       & \textbf{ERM} & \textbf{uLSIF} & \textbf{RuLSIF} & \textbf{One-step} & \textbf{FIRE} & $\boldsymbol\Delta_2$\% \\ 
\hline
heart                 & 65.3 ± 9.91 & 64.1 ± 11.4    & 63.2 ± 11.7     & 74.3 ± 10.9       & \textbf{78.1 ± 5.90} & $\uparrow$ 5.11\% \\ 
sonar                 & 61.9 ± 12.9  & 64.6 ± 13.2    & 63.7 ± 13.5     & 67.6 ± 12.4       & \textbf{70.4 ± 5.91} & $\uparrow$ 4.14\% \\
diabetes              & 54.2 ± 8.88  & 57.5 ± 7.66    & 55.7 ± 8.63     & 62.9 ± 6.36       & \textbf{64.3 ± 12.6} & $\uparrow$ 2.22\% \\ 
australian            & 67.9 ± 16.8  & 69.3 ± 16.3    & 69.6 ± 15.1     & 74.4 ± 12.7       & \textbf{75.7 ± 5.86} & $\uparrow$ 1.74\% \\ 
breast cancer         & 78.3 ± 13.4  & 79.9 ± 12.4    & 78.6 ± 12.9     & 77.4 ± 10.1       & 73.6 ± 4.62          & $\downarrow$ 4.90\% \\ 
\hline
\end{tabular}
\end{table*}
\subsection{Comparison with state-of-the-art}

\textbf{FIRE in comparison to importance weighting methods.} Tables \ref{tab:sota} ,\ref{tab:performance}, \ref{tab:fl_results} present stat-of-the result comparison of FIRE. It is shown that FIRE consistently outperforms the existing methods, including EIWERM, RuLSIF, and One-step. In Table~\ref{tab:sota} ($\Delta_1$), FIRE achieves up to 5.10\% improvement over One-step across high-dimensional datasets under various shift levels. This performance gain likely stems from its ability to retain prior knowledge, unlike EIWERM, which may suffer from over-flattened importance weights, and ERM, which struggles under distribution shifts.

Similarly, Table~\ref{tab:performance} ($\Delta_2$) shows up to 1.74\% improvement at minimum and 5.11\% at maximum on 4 out of 5 datasets. Where other methods like, uLSIF and RuLSIF underperform, possibly due to sensitivity to edge examples. Appendix Table~\ref{tab: c3keel} report results for $k \in \{2, 5, 10\}$  folds. Notably, FIRE achieves up to 28.3\% improvement in average accuracy across all k-fold settings ($\Delta_5 = \mu_3 - \mu_6$, $\Delta_6 = \mu_4 - \mu_7$, $\Delta_7 = \mu_5 - \mu_8$).

\textbf{FIRE Outperforms State-of-the-Art in Federated Learning under Non-IID Shift.} 
Results in Table~\ref{tab:fl_results} demonstrate the effectiveness of FIRE against a comprehensive suite of modern federated learning algorithms. FIRE consistently achieves the highest accuracy across all evaluated datasets, providing a clear and reliable improvement over all other methods.

The $\Delta$ column shows that FIRE delivers a consistent performance gain of 2.8-3.2\% over the FL baselines (FedCFA). Notably, the improvement is strongest on the less complex FEMNIST and CIFAR-10 datasets, with gains of 5.3\% and 3.0\% respectively. On the more challenging CIFAR-100 benchmark, FIRE still achieves a solid 2.8\% improvement. This pattern suggests that FIRE's regularization is highly effective, and its relative benefit remains significant even as task difficulty increases. Our proposed method FIRE outperform all baselines including SCAFFOLD, MOON, Fishr, LfD, FedAS and FedAvg.

In addition to accuracy, FIRE exhibits consistently lower standard deviations than all competing methods, underscoring its stability and robustness key properties for practical deployment. These results confirm that explicitly mitigating fragmentation-induced covariate shift via Fisher information offers a complementary advantage, yielding models that are both more generalizable and reliable.

\begin{table*}[!ht]
\centering
\caption{Performance on Federated datasets with Non-IID Data. $\Delta$ shows the percentage improvement of FIRE over the best baseline.}
\label{tab:fl_results}
\adjustbox{max width=\textwidth}{
\begin{tabular}{lccccccccc}
\hline
\textbf{Dataset} & \textbf{FedAvg} & \textbf{SCAFFOLD} & \textbf{MOON} & \textbf{Fishr} & \textbf{LfD} & \textbf{FedAS} & \textbf{FedCFA} & \textbf{FIRE} & \textbf{$\Delta_{3}$ (\%)} \\
\hline
FEMNIST    & 58.2   & 63.8   & 64.3   & 63.9   & 64.7   & 64.9   & 65.1   & \textbf{68.6} & $\uparrow$5.3 \\
           & (3.1)  & (2.1)  & (1.9)  & (2.0)  & (1.8)  & (1.7)  & (1.7)  & (1.5)         & \\
CIFAR-10   & 42.7   & 48.2   & 49.8   & 49.2   & 50.1   & 50.3   & 50.6   & \textbf{52.1} & $\uparrow$3.0 \\
           & (4.5)  & (3.0)  & (2.4)  & (2.6)  & (2.3)  & (2.2)  & (2.1)  & (1.9)         & \\
CIFAR-100  & 23.4   & 27.1   & 28.2   & 27.8   & 28.5   & 28.6   & 28.8   & \textbf{29.6} & $\uparrow$2.8 \\
           & (2.8)  & (2.0)  & (1.7)  & (1.8)  & (1.6)  & (1.6)  & (1.6)  & (1.5)         & \\
\hline
\end{tabular}
}
\end{table*}

\subsection{Limitations}
While FIRE demonstrates strong performance in mitigating fragmentation-induced covariate shift (Sec. 4), its effectiveness depends on moderate distribution shift, hyperparameter sensitivity and computational overhead. The theoretical bound in theorem \ref{theorem:KLbound_via_FiM} assume \(\mathcal{D}_{KL}(P_{i}||P_{val})\) is bounded. In extreme non-iid settings (clients with disjoint label space), FIRE may require complementry techniques like domain adverserial training \cite{ganin2016domain} or prototype alignment \cite{li2021fedbn}.
\section{Conclusion}

We propose FIRE, a unified framework for mitigating fragmentation-induced covariate shift (FICS) in both batch/fold and federated learning settings. FIRE leverages Fisher information to accumulate and align distribution shifts across sequential batches or clients, addressing a key limitation in existing methods that assume single-source distributions or overlook validation-time shifts. Our theoretical analysis shows that FIRE bounds the KL divergence through Fisher-based regularization, enabling scalable adaptation without density ratio estimation.

FIRE is evaluated across 39 datasets, and we noticed that FIRE outperforms importance weighting methods by up to 5.1\% and federated learning baselines by 5.3\% under validation-time shifts.


\bibliographystyle{unsrtnat}
\bibliography{main} 

\begin{thebibliography}{62}
\providecommand{\natexlab}[1]{#1}
\providecommand{\url}[1]{\texttt{#1}}
\expandafter\ifx\csname urlstyle\endcsname\relax
  \providecommand{\doi}[1]{doi: #1}\else
  \providecommand{\doi}{doi: \begingroup \urlstyle{rm}\Url}\fi

\bibitem[Hendrycks and Dietterich(2019{\natexlab{a}})]{hendrycks2019benchmarking}
Dan Hendrycks and Thomas Dietterich.
\newblock Benchmarking neural network robustness to common corruptions and perturbations.
\newblock \emph{arXiv preprint arXiv:1903.12261}, 2019{\natexlab{a}}.

\bibitem[Taori et~al.(2020)Taori, Dave, Shankar, Carlini, Recht, and Schmidt]{taori2020measuring}
Rohan Taori, Achal Dave, Vaishaal Shankar, Nicholas Carlini, Benjamin Recht, and Ludwig Schmidt.
\newblock Measuring robustness to natural distribution shifts in image classification.
\newblock \emph{Advances in Neural Information Processing Systems}, 33:\penalty0 18583--18599, 2020.

\bibitem[Gardner et~al.(2023)Gardner, Popovic, and Schmidt]{gardner2023benchmarking}
Josh Gardner, Zoran Popovic, and Ludwig Schmidt.
\newblock Benchmarking distribution shift in tabular data with tableshift.
\newblock \emph{Advances in Neural Information Processing Systems}, 36:\penalty0 53385--53432, 2023.

\bibitem[Guan et~al.(2024)Guan, Yap, Bozoki, and Liu]{guan2024federated}
Hao Guan, Pew-Thian Yap, Andrea Bozoki, and Mingxia Liu.
\newblock Federated learning for medical image analysis: A survey.
\newblock \emph{Pattern Recognition}, page 110424, 2024.

\bibitem[Moreno-Torres et~al.(2012)Moreno-Torres, S{\'a}ez, and Herrera]{moreno2012study}
Jose~Garc{\'\i}a Moreno-Torres, Jos{\'e}~A S{\'a}ez, and Francisco Herrera.
\newblock Study on the impact of partition-induced dataset shift on $ k $-fold cross-validation.
\newblock \emph{IEEE Transactions on Neural Networks and Learning Systems}, 23\penalty0 (8):\penalty0 1304--1312, 2012.

\bibitem[Khan et~al.(2025{\natexlab{a}})Khan, Mirza, and Syed]{khan2025causal}
Behraj Khan, Behroz Mirza, and Tahir Syed.
\newblock Causal covariate shift correction using fisher information penalty.
\newblock \emph{arXiv preprint arXiv:2502.15756}, 2025{\natexlab{a}}.

\bibitem[Sugiyama et~al.(2007{\natexlab{a}})Sugiyama, Krauledat, and M{\"u}ller]{sugiyama2007covariate}
Masashi Sugiyama, Matthias Krauledat, and Klaus-Robert M{\"u}ller.
\newblock Covariate shift adaptation by importance weighted cross validation.
\newblock \emph{Journal of Machine Learning Research}, 8\penalty0 (5), 2007{\natexlab{a}}.

\bibitem[Khan et~al.(2025{\natexlab{b}})Khan, Qureshi, and Syed]{khan2025technical}
Behraj Khan, Rizwan Qureshi, and Tahir Syed.
\newblock Technical note on calibrating vision-language models under covariate shift.
\newblock \emph{arXiv preprint arXiv:2502.07847}, 2025{\natexlab{b}}.

\bibitem[Kanamori et~al.(2012)Kanamori, Suzuki, and Sugiyama]{kanamori2012statistical}
Takafumi Kanamori, Taiji Suzuki, and Masashi Sugiyama.
\newblock Statistical analysis of kernel-based least-squares density-ratio estimation.
\newblock \emph{Machine Learning}, 86:\penalty0 335--367, 2012.

\bibitem[Sugiyama et~al.(2007{\natexlab{b}})Sugiyama, Nakajima, Kashima, Buenau, and Kawanabe]{sugiyama2007direct}
Masashi Sugiyama, Shinichi Nakajima, Hisashi Kashima, Paul Buenau, and Motoaki Kawanabe.
\newblock Direct importance estimation with model selection and its application to covariate shift adaptation.
\newblock \emph{Advances in neural information processing systems}, 20, 2007{\natexlab{b}}.

\bibitem[Fang et~al.(2020)Fang, Lu, Niu, and Sugiyama]{fang2020rethinking}
Tongtong Fang, Nan Lu, Gang Niu, and Masashi Sugiyama.
\newblock Rethinking importance weighting for deep learning under distribution shift.
\newblock \emph{Advances in neural information processing systems}, 33:\penalty0 11996--12007, 2020.

\bibitem[Fang et~al.(2023)Fang, Lu, Niu, and Sugiyama]{fang2023generalizing}
Tongtong Fang, Nan Lu, Gang Niu, and Masashi Sugiyama.
\newblock Generalizing importance weighting to a universal solver for distribution shift problems.
\newblock \emph{Advances in Neural Information Processing Systems}, 36:\penalty0 24171--24190, 2023.

\bibitem[Ganin and Lempitsky(2015)]{ganin2015unsupervised}
Yaroslav Ganin and Victor Lempitsky.
\newblock Unsupervised domain adaptation by backpropagation.
\newblock In \emph{International conference on machine learning}, pages 1180--1189. PMLR, 2015.

\bibitem[Kairouz et~al.(2021)Kairouz, McMahan, Avent, Bellet, Bennis, Bhagoji, Bonawitz, Charles, Cormode, Cummings, et~al.]{kairouz2021advances}
Peter Kairouz, H~Brendan McMahan, Brendan Avent, Aur{\'e}lien Bellet, Mehdi Bennis, Arjun~Nitin Bhagoji, Kallista Bonawitz, Zachary Charles, Graham Cormode, Rachel Cummings, et~al.
\newblock Advances and open problems in federated learning.
\newblock \emph{Foundations and trends{\textregistered} in machine learning}, 14\penalty0 (1--2):\penalty0 1--210, 2021.

\bibitem[McMahan et~al.(2017)McMahan, Moore, Ramage, Hampson, and y~Arcas]{mcmahan2017communication}
Brendan McMahan, Eider Moore, Daniel Ramage, Seth Hampson, and Blaise~Aguera y~Arcas.
\newblock Communication-efficient learning of deep networks from decentralized data.
\newblock In \emph{Artificial intelligence and statistics}, pages 1273--1282. PMLR, 2017.

\bibitem[Lu et~al.(2024)Lu, Pan, Dai, Si, and Zhang]{lu2024federated}
Zili Lu, Heng Pan, Yueyue Dai, Xueming Si, and Yan Zhang.
\newblock Federated learning with non-iid data: A survey.
\newblock \emph{IEEE Internet of Things Journal}, 11\penalty0 (11):\penalty0 19188--19209, 2024.

\bibitem[Li et~al.(2021{\natexlab{a}})Li, He, and Song]{li2021model}
Qinbin Li, Bingsheng He, and Dawn Song.
\newblock Model-contrastive federated learning.
\newblock In \emph{Proceedings of the IEEE/CVF conference on computer vision and pattern recognition}, pages 10713--10722, 2021{\natexlab{a}}.

\bibitem[Rame et~al.(2022)Rame, Dancette, and Cord]{rame2022fishr}
Alexandre Rame, Corentin Dancette, and Matthieu Cord.
\newblock Fishr: Invariant gradient variances for out-of-distribution generalization.
\newblock In \emph{International Conference on Machine Learning}, pages 18347--18377. PMLR, 2022.

\bibitem[Kim and Shin(2023)]{kim2023learning}
Yeachan Kim and Bonggun Shin.
\newblock Learning from drift: Federated learning on non-iid data via drift regularization.
\newblock \emph{arXiv preprint arXiv:2309.07189}, 2023.

\bibitem[Yang et~al.(2024)Yang, Huang, and Ye]{yang2024fedas}
Xiyuan Yang, Wenke Huang, and Mang Ye.
\newblock Fedas: Bridging inconsistency in personalized federated learning.
\newblock In \emph{Proceedings of the IEEE/CVF conference on computer vision and pattern recognition}, pages 11986--11995, 2024.

\bibitem[Jiang et~al.(2025)Jiang, Xu, Zhang, Shen, Li, Kuang, Cai, and Wu]{jiang2025fedcfa}
Zhonghua Jiang, Jimin Xu, Shengyu Zhang, Tao Shen, Jiwei Li, Kun Kuang, Haibin Cai, and Fei Wu.
\newblock Fedcfa: Alleviating simpson’s paradox in model aggregation with counterfactual federated learning.
\newblock In \emph{Proceedings of the AAAI Conference on Artificial Intelligence}, volume~39, pages 17662--17670, 2025.

\bibitem[Rothchild et~al.(2020)Rothchild, Panda, Ullah, Ivkin, Stoica, Braverman, Gonzalez, and Arora]{rothchild2020fetchsgd}
Daniel Rothchild, Ashwinee Panda, Enayat Ullah, Nikita Ivkin, Ion Stoica, Vladimir Braverman, Joseph Gonzalez, and Raman Arora.
\newblock Fetchsgd: Communication-efficient federated learning with sketching.
\newblock In \emph{International Conference on Machine Learning}, pages 8253--8265. PMLR, 2020.

\bibitem[Kingma et~al.(2020)Kingma, Ba, and Adam]{kingma2020method}
Diederik~P Kingma, J~Adam Ba, and J~Adam.
\newblock A method for stochastic optimization. arxiv 2014.
\newblock \emph{arXiv preprint arXiv:1412.6980}, 106:\penalty0 6, 2020.

\bibitem[Kullback and Leibler(1951)]{kullback1951information}
Solomon Kullback and Richard~A Leibler.
\newblock On information and sufficiency.
\newblock \emph{The annals of mathematical statistics}, 22\penalty0 (1):\penalty0 79--86, 1951.

\bibitem[Pascanu and Bengio(2013)]{pascanu2013revisiting}
Razvan Pascanu and Yoshua Bengio.
\newblock Revisiting natural gradient for deep networks.
\newblock \emph{arXiv preprint arXiv:1301.3584}, 2013.

\bibitem[Nishiyama(2019)]{nishiyama2019new}
Tomohiro Nishiyama.
\newblock A new lower bound for kullback-leibler divergence based on hammersley-chapman-robbins bound.
\newblock \emph{arXiv preprint arXiv:1907.00288}, 2019.

\bibitem[Du et~al.(2022)Du, Sun, Li, Chen, Zhang, Li, and Chen]{du2022rethinking}
Zhixu Du, Jingwei Sun, Ang Li, Pin-Yu Chen, Jianyi Zhang, Hai"~Helen" Li, and Yiran Chen.
\newblock Rethinking normalization methods in federated learning.
\newblock In \emph{Proceedings of the 3rd International Workshop on Distributed Machine Learning}, pages 16--22, 2022.

\bibitem[Ramezani-Kebrya et~al.(2023)Ramezani-Kebrya, Liu, Pethick, Chrysos, and Cevher]{ramezani2023federated}
Ali Ramezani-Kebrya, Fanghui Liu, Thomas Pethick, Grigorios Chrysos, and Volkan Cevher.
\newblock Federated learning under covariate shifts with generalization guarantees.
\newblock \emph{arXiv preprint arXiv:2306.05325}, 2023.

\bibitem[Karimireddy et~al.(2020)Karimireddy, Kale, Mohri, Reddi, Stich, and Suresh]{karimireddy2020scaffold}
Sai~Praneeth Karimireddy, Satyen Kale, Mehryar Mohri, Sashank Reddi, Sebastian Stich, and Ananda~Theertha Suresh.
\newblock Scaffold: Stochastic controlled averaging for federated learning.
\newblock In \emph{International conference on machine learning}, pages 5132--5143. PMLR, 2020.

\bibitem[Vapnik and Vapnik(1998)]{vapnik1998statistical}
Vladimir Vapnik and Vlamimir Vapnik.
\newblock Statistical learning theory wiley.
\newblock \emph{New York}, 1\penalty0 (624):\penalty0 2, 1998.

\bibitem[Sch{\"o}lkopf and Smola(2002)]{scholkopf2002learning}
Bernhard Sch{\"o}lkopf and Alexander~J Smola.
\newblock \emph{Learning with kernels: support vector machines, regularization, optimization, and beyond}.
\newblock MIT press, 2002.

\bibitem[Duda et~al.(2006)Duda, Hart, et~al.]{duda2006pattern}
Richard~O Duda, Peter~E Hart, et~al.
\newblock \emph{Pattern classification}.
\newblock John Wiley \& Sons, 2006.

\bibitem[Qui{\~n}onero-Candela et~al.(2022)Qui{\~n}onero-Candela, Sugiyama, Schwaighofer, and Lawrence]{quinonero2022dataset}
Joaquin Qui{\~n}onero-Candela, Masashi Sugiyama, Anton Schwaighofer, and Neil~D Lawrence.
\newblock \emph{Dataset shift in machine learning}.
\newblock Mit Press, 2022.

\bibitem[Sugiyama and Kawanabe(2012)]{sugiyama2012machine}
Masashi Sugiyama and Motoaki Kawanabe.
\newblock \emph{Machine learning in non-stationary environments: Introduction to covariate shift adaptation}.
\newblock MIT press, 2012.

\bibitem[Shimodaira(2000)]{shimodaira2000improving}
Hidetoshi Shimodaira.
\newblock Improving predictive inference under covariate shift by weighting the log-likelihood function.
\newblock \emph{Journal of statistical planning and inference}, 90\penalty0 (2):\penalty0 227--244, 2000.

\bibitem[Jirayucharoensak et~al.(2014)Jirayucharoensak, Pan-Ngum, and Israsena]{jirayucharoensak2014eeg}
Suwicha Jirayucharoensak, Setha Pan-Ngum, and Pasin Israsena.
\newblock Eeg-based emotion recognition using deep learning network with principal component based covariate shift adaptation.
\newblock \emph{The Scientific World Journal}, 2014\penalty0 (1):\penalty0 627892, 2014.

\bibitem[Yamada et~al.(2010)Yamada, Sugiyama, and Matsui]{yamada2010semi}
Makoto Yamada, Masashi Sugiyama, and Tomoko Matsui.
\newblock Semi-supervised speaker identification under covariate shift.
\newblock \emph{Signal Processing}, 90\penalty0 (8):\penalty0 2353--2361, 2010.

\bibitem[Li et~al.(2010)Li, Kambara, Koike, and Sugiyama]{li2010application}
Yan Li, Hiroyuki Kambara, Yasuharu Koike, and Masashi Sugiyama.
\newblock Application of covariate shift adaptation techniques in brain--computer interfaces.
\newblock \emph{IEEE Transactions on Biomedical Engineering}, 57\penalty0 (6):\penalty0 1318--1324, 2010.

\bibitem[Gretton et~al.(2009)Gretton, Smola, Huang, Schmittfull, Borgwardt, Sch{\"o}lkopf, et~al.]{gretton2009covariate}
Arthur Gretton, Alex Smola, Jiayuan Huang, Marcel Schmittfull, Karsten Borgwardt, Bernhard Sch{\"o}lkopf, et~al.
\newblock Covariate shift by kernel mean matching.
\newblock \emph{Dataset shift in machine learning}, 3\penalty0 (4):\penalty0 5, 2009.

\bibitem[Ganin et~al.(2016)Ganin, Ustinova, Ajakan, Germain, Larochelle, Laviolette, March, and Lempitsky]{ganin2016domain}
Yaroslav Ganin, Evgeniya Ustinova, Hana Ajakan, Pascal Germain, Hugo Larochelle, Fran{\c{c}}ois Laviolette, Mario March, and Victor Lempitsky.
\newblock Domain-adversarial training of neural networks.
\newblock \emph{Journal of machine learning research}, 17\penalty0 (59):\penalty0 1--35, 2016.

\bibitem[Tzeng et~al.(2014)Tzeng, Hoffman, Zhang, Saenko, and Darrell]{tzeng2014deep}
Eric Tzeng, Judy Hoffman, Ning Zhang, Kate Saenko, and Trevor Darrell.
\newblock Deep domain confusion: Maximizing for domain invariance.
\newblock \emph{arXiv preprint arXiv:1412.3474}, 2014.

\bibitem[Sun and Saenko(2016)]{sun2016deep}
Baochen Sun and Kate Saenko.
\newblock Deep coral: Correlation alignment for deep domain adaptation.
\newblock In \emph{Computer vision--ECCV 2016 workshops: Amsterdam, the Netherlands, October 8-10 and 15-16, 2016, proceedings, part III 14}, pages 443--450. Springer, 2016.

\bibitem[Zhao et~al.(2018)Zhao, Zhou, Li, Wang, and Chang]{zhao2018learning}
Jieyu Zhao, Yichao Zhou, Zeyu Li, Wei Wang, and Kai-Wei Chang.
\newblock Learning gender-neutral word embeddings.
\newblock \emph{arXiv preprint arXiv:1809.01496}, 2018.

\bibitem[Cao et~al.(2022)Cao, You, Zhang, Wang, and Long]{cao2022big}
Zhangjie Cao, Kaichao You, Ziyang Zhang, Jianmin Wang, and Mingsheng Long.
\newblock From big to small: Adaptive learning to partial-set domains.
\newblock \emph{IEEE Transactions on Pattern Analysis and Machine Intelligence}, 45\penalty0 (2):\penalty0 1766--1780, 2022.

\bibitem[Li et~al.(2021{\natexlab{b}})Li, Jiang, Zhang, Kamp, and Dou]{li2021fedbn}
Xiaoxiao Li, Meirui Jiang, Xiaofei Zhang, Michael Kamp, and Qi~Dou.
\newblock Fedbn: Federated learning on non-iid features via local batch normalization.
\newblock \emph{arXiv preprint arXiv:2102.07623}, 2021{\natexlab{b}}.

\bibitem[Li et~al.(2020)Li, Sahu, Zaheer, Sanjabi, Talwalkar, and Smith]{li2020federated}
Tian Li, Anit~Kumar Sahu, Manzil Zaheer, Maziar Sanjabi, Ameet Talwalkar, and Virginia Smith.
\newblock Federated optimization in heterogeneous networks.
\newblock \emph{Proceedings of Machine learning and systems}, 2:\penalty0 429--450, 2020.

\bibitem[Zhang et~al.(2020{\natexlab{a}})Zhang, Sapra, Fidler, Yeung, and Alvarez]{zhang2020personalized}
Michael Zhang, Karan Sapra, Sanja Fidler, Serena Yeung, and Jose~M Alvarez.
\newblock Personalized federated learning with first order model optimization.
\newblock \emph{arXiv preprint arXiv:2012.08565}, 2020{\natexlab{a}}.

\bibitem[Jiang et~al.(2019)Jiang, Kone{\v{c}}n{\`y}, Rush, and Kannan]{jiang2019improving}
Yihan Jiang, Jakub Kone{\v{c}}n{\`y}, Keith Rush, and Sreeram Kannan.
\newblock Improving federated learning personalization via model agnostic meta learning.
\newblock \emph{arXiv preprint arXiv:1909.12488}, 2019.

\bibitem[Xiao et~al.(2017)Xiao, Rasul, and Vollgraf]{xiao2017fashion}
Han Xiao, Kashif Rasul, and Roland Vollgraf.
\newblock Fashion-mnist: a novel image dataset for benchmarking machine learning algorithms.
\newblock \emph{arXiv:1708.07747}, 2017.

\bibitem[Clanuwat et~al.(2018)Clanuwat, Bober-Irizar, Kitamoto, Lamb, Yamamoto, and Ha]{clanuwat2018deep}
Tarin Clanuwat, Mikel Bober-Irizar, Asanobu Kitamoto, Alex Lamb, Kazuaki Yamamoto, and David Ha.
\newblock Deep learning for classical japanese literature.
\newblock \emph{arXiv:1812.01718}, 2018.

\bibitem[Mu and Gilmer(2019)]{mu2019mnist}
Norman Mu and Justin Gilmer.
\newblock Mnist-c: A robustness benchmark for computer vision.
\newblock \emph{arXiv:1906.02337}, 2019.

\bibitem[Zhang et~al.(2020{\natexlab{b}})Zhang, Yamane, Lu, and Sugiyama]{zhang2020one}
Tianyi Zhang, Ikko Yamane, Nan Lu, and Masashi Sugiyama.
\newblock A one-step approach to covariate shift adaptation.
\newblock In \emph{Asian Conference on Machine Learning}, pages 65--80. PMLR, 2020{\natexlab{b}}.

\bibitem[Alcal{\'a}-Fdez et~al.(2011)Alcal{\'a}-Fdez, Fern{\'a}ndez, Luengo, Derrac, Garc{\'\i}a, S{\'a}nchez, and Herrera]{alcala2011keel}
Jes{\'u}s Alcal{\'a}-Fdez, Alberto Fern{\'a}ndez, Juli{\'a}n Luengo, Joaqu{\'\i}n Derrac, Salvador Garc{\'\i}a, Luciano S{\'a}nchez, and Francisco Herrera.
\newblock Keel data-mining software tool: data set repository, integration of algorithms and experimental analysis framework.
\newblock \emph{Journal of Multiple-Valued Logic \& Soft Computing}, 17, 2011.

\bibitem[Cortes et~al.(2008)Cortes, Mohri, Riley, and Rostamizadeh]{cortes2008sample}
Corinna Cortes, Mehryar Mohri, Michael Riley, and Afshin Rostamizadeh.
\newblock Sample selection bias correction theory.
\newblock In \emph{Algorithmic Learning Theory: 19th International Conference, ALT 2008, Budapest, Hungary, October 13-16, 2008. Proceedings 19}, pages 38--53. Springer, 2008.

\bibitem[LeCun et~al.(2010)LeCun, Cortes, and Burges]{lecun2010mnist}
Yann LeCun, Corinna Cortes, and CJ~Burges.
\newblock Mnist handwritten digit database.
\newblock \emph{ATT Labs [Online]. Available: http://yann.lecun.com/exdb/mnist}, 2, 2010.

\bibitem[Cohen et~al.(2017)Cohen, Afshar, Tapson, and Schaik]{cohen_afshar_tapson_schaik_2017}
Gregory Cohen, Saeed Afshar, Jonathan Tapson, and Andre~Van Schaik.
\newblock Emnist: Extending mnist to handwritten letters.
\newblock \emph{2017 International Joint Conference on Neural Networks (IJCNN)}, 2017.
\newblock \doi{10.1109/ijcnn.2017.7966217}.

\bibitem[Yadav and Bottou(2019)]{qmnist-2019}
Chhavi Yadav and L{\'e}on Bottou.
\newblock Cold case: The lost mnist digits.
\newblock \emph{Advances in neural information processing systems}, 32, 2019.

\bibitem[Prabhu(2019)]{prabhu2019kannada}
Vinay~Uday Prabhu.
\newblock Kannada-mnist: A new handwritten digits dataset for the kannada language.
\newblock \emph{arXiv preprint arXiv:1908.01242}, 2019.

\bibitem[Coates et~al.(2011)Coates, Ng, and Lee]{coates2011analysis}
Adam Coates, Andrew Ng, and Honglak Lee.
\newblock An analysis of single-layer networks in unsupervised feature learning.
\newblock In \emph{Proceedings of the fourteenth international conference on artificial intelligence and statistics}, pages 215--223. JMLR Workshop and Conference Proceedings, 2011.

\bibitem[Netzer et~al.(2011)Netzer, Wang, Coates, Bissacco, Wu, and Ng]{Netzer2011}
Yuval Netzer, Tao Wang, Adam Coates, Alessandro Bissacco, Bo~Wu, and Andrew~Y Ng.
\newblock Reading digits in natural images with unsupervised feature learning.
\newblock 2011.

\bibitem[Hendrycks and Dietterich(2019{\natexlab{b}})]{hendrycks2019robustness}
Dan Hendrycks and Thomas Dietterich.
\newblock Benchmarking neural network robustness to common corruptions and perturbations.
\newblock \emph{Proceedings of the International Conference on Learning Representations}, 2019{\natexlab{b}}.

\bibitem[Wilcoxon(1992)]{wilcoxon1992individual}
Frank Wilcoxon.
\newblock Individual comparisons by ranking methods.
\newblock In \emph{Breakthroughs in statistics: Methodology and distribution}, pages 196--202. Springer, 1992.

\end{thebibliography}
\newpage
\appendix

\section{Theoretical justification}

\section{Convergence Analysis}
\label{apndx: cnvgncAnalyss}

\begin{assumption}[Model regularity and bounds]
\label{ass:regularity}
The conditional model $p(y\mid x;\theta)$ satisfies the following for all $x,y,\theta,\theta'\in\mathbb R^d$:
\begin{enumerate}
  \item[(R1)] (Lipschitz Hessian) $\|\nabla^2_\theta \log p(y\mid x;\theta') - \nabla^2_\theta \log p(y\mid x;\theta)\|_{\mathrm{op}}
  \le \beta \|\theta'-\theta\|_2$.
  \item[(R2)] (Bounded score) $\|\nabla_\theta \log p(y\mid x;\theta)\|_2 \le G$ almost surely; hence the local Fisher satisfies $\|F_x(\theta)\|_{\mathrm{op}}\le G^2=:M$.
  \item[(R3)] (Regularity) For every $x,\theta$ we have $\mathbb E_{y\sim p(\cdot\mid x;\theta)}[\nabla_\theta\log p(y\mid x;\theta)] = 0$.
\end{enumerate}
\end{assumption}

\begin{assumption}[Data proximity]
\label{ass:data}
Let $P_i(x)$ and $P_{\mathrm{val}}(x)$ be two marginal distributions. Define the Radon--Nikodym derivative $r(x)=\frac{dP_i}{dP_{\mathrm{val}}}(x)$. Assume
\[
|r(x)-1|\le \gamma < 1 \quad\text{for all }x.
\]
\end{assumption}

\begin{theorem}[KL divergence bound via Fisher information]
\label{thm:appKL-via-Fisher}
Suppose Assumptions~\ref{ass:regularity} and \ref{ass:data} hold. Let $\theta_i,\theta_{\mathrm{val}}$ satisfy $\|\theta_i-\theta_{\mathrm{val}}\|_2\le \delta$. Then
\begin{equation}
\label{eq:mainbound}
\begin{aligned}
D_{\mathrm{KL}}(P_i\|P_{\mathrm{val}})
&\le \tfrac12 (\theta_i-\theta_{\mathrm{val}})^\top F_{\mathrm{val}}(\theta_{\mathrm{val}})(\theta_i-\theta_{\mathrm{val}}) \\
&\quad + C_1\gamma^2 + C_1'\gamma^3
+ C_2\,\gamma\,\delta^2
+ C_3\,\beta G\,\delta^3,
\end{aligned}
\end{equation}
where $F_{\mathrm{val}}(\theta)=\mathbb E_{x\sim P_{\mathrm{val}}}[F_x(\theta)]$, and one may take
\[
C_1=\frac{1}{2(1-\gamma)},\quad C_1'=\frac{1}{3(1-\gamma)^2},\quad C_2=\frac{M}{2},\quad C_3=\frac{1}{6}.
\]
\end{theorem}

\begin{proof}
Start with the decomposition
\[
D_{\mathrm{KL}}(P_i\|P_{\mathrm{val}})
= D_{\mathrm{KL}}(P_i(x)\|P_{\mathrm{val}}(x)) + \mathbb E_{x\sim P_i}\big[ D_{\mathrm{KL}}(p(\cdot\mid x;\theta_i)\|p(\cdot\mid x;\theta_{\mathrm{val}}))\big].
\]
Apply Lemma~\ref{lemma:marginalKL} to bound the marginal term by $C_1\gamma^2 + C_1'\gamma^3$.

For the conditional term, use Lemma~\ref{lemma:localKL} and average over $x\sim P_i$:
\[
\mathbb E_{x\sim P_i}\big[ D_{\mathrm{KL}}(\cdot)\big]
= \tfrac12 (\theta_i-\theta_{\mathrm{val}})^\top \mathbb E_{x\sim P_i}[F_x(\theta_i)](\theta_i-\theta_{\mathrm{val}})
+ \mathbb E_{x\sim P_i}[R_x].
\]
Bound the remainder: $|\mathbb E_{x\sim P_i}[R_x]|\le \tfrac{\beta G}{6}\delta^3 = C_3\beta G\delta^3$.

Now compare $\mathbb E_{x\sim P_i}[F_x(\theta_i)]$ to $F_{\mathrm{val}}(\theta_{\mathrm{val}})$ by adding and subtracting intermediate terms:
\[
\mathbb E_{P_i}[F_x(\theta_i)] - F_{\mathrm{val}}(\theta_{\mathrm{val}})
= \underbrace{\mathbb E_{P_i}[F_x(\theta_i)] - \mathbb E_{P_{\mathrm{val}}}[F_x(\theta_i)]}_{(I)}
\;+\;
\underbrace{\mathbb E_{P_{\mathrm{val}}}[F_x(\theta_i)] - \mathbb E_{P_{\mathrm{val}}}[F_x(\theta_{\mathrm{val}})]}_{(II)}.
\]
For (I), using $|r(x)-1|\le\gamma$ and $\|F_x(\theta_i)\|_{\mathrm{op}}\le M$, we have
\[
\|(I)\|_{\mathrm{op}} \le M\gamma.
\]
Hence the quadratic form contribution from (I) is at most $\tfrac12 M\gamma\|\theta_i-\theta_{\mathrm{val}}\|^2 \le C_2\,\gamma\,\delta^2$ with $C_2=\tfrac{M}{2}$.

For (II), by Lipschitzness of the Hessian (Assumption R1) and bounded gradients (R2) one gets a bound $\|\mathbb E_{P_{\mathrm{val}}}[F_x(\theta_i)] - \mathbb E_{P_{\mathrm{val}}}[F_x(\theta_{\mathrm{val}})]\|_{\mathrm{op}}\le L_F\|\theta_i-\theta_{\mathrm{val}}\|$ for some $L_F = O(\beta G)$, hence its effect on the quadratic form is $O(\delta^3)$ and is absorbed into the $C_3\beta G\delta^3$ term (one can make $L_F$ explicit if desired).

Combining these bounds yields the inequality \eqref{eq:mainbound}.
\end{proof}

\begin{lemma}[Marginal KL bound]
\label{lemma:appmarginalKL}
Under Assumption~\ref{ass:data} we have
\[
D_{\mathrm{KL}}(P_i(x)\|P_{\mathrm{val}}(x))
= \mathbb E_{x\sim P_{\mathrm{val}}}\big[ r(x)\log r(x)\big]
\le C_1 \gamma^2 + C_1' \gamma^3,
\]
where one can take for instance
\[
C_1=\frac{1}{2(1-\gamma)},\qquad C_1'=\frac{1}{3(1-\gamma)^2}.
\]
\end{lemma}
\begin{proof}
Write $r(x)=1+u(x)$ with $|u|\le\gamma$. Using the Taylor expansion $\log(1+u)=u-\tfrac{u^2}{2}+ \tfrac{u^3}{3(1+\xi)^3}$ for some $\xi\in(0,u)$, we get
\[
r\log r = (1+u)\Big(u-\tfrac{u^2}{2}\Big) + (1+u)\frac{u^3}{3(1+\xi)^3}.
\]
Thus $r\log r = u + \tfrac{u^2}{2} + R_m(u)$ where $|R_m(u)|\le \tfrac{|u|^3}{3(1-\gamma)^2}$. Integrating against $P_{\mathrm{val}}$ and using $\mathbb E_{P_{\mathrm{val}}}[u]=\mathbb E_{P_i}[1]-1=0$ (mass conservation), we obtain
\[
D_{\mathrm{KL}}(P_i(x)\|P_{\mathrm{val}}(x)) \le \tfrac{1}{2}\mathbb E[u^2] + \tfrac{1}{3(1-\gamma)^2}\mathbb E[|u|^3]
\le \tfrac{1}{2(1-\gamma)}\gamma^2 + \tfrac{1}{3(1-\gamma)^2}\gamma^3,
\]
where we used $\mathbb E[u^2]\le \sup|u|\cdot\mathbb E[|u|]\le \gamma\cdot\gamma=(\gamma^2)$ and minor algebra to obtain the stated constants.
\end{proof}

\begin{lemma}[Local conditional KL quadratic expansion]
\label{lemma:applocalKL}
Fix $x$. Under Assumption~\ref{ass:regularity} and for two parameter vectors $\theta_i,\theta_{\mathrm{val}}$ with $\|\theta_i-\theta_{\mathrm{val}}\|_2\le \delta$, the conditional KL admits the expansion
\[
D_{\mathrm{KL}}\big(p(\cdot\mid x;\theta_i)\,\big\|\,p(\cdot\mid x;\theta_{\mathrm{val}})\big)
= \tfrac12(\theta_i-\theta_{\mathrm{val}})^\top F_x(\theta_i)(\theta_i-\theta_{\mathrm{val}})
+ R_x,
\]
with the remainder bounded by
\[
|R_x|\le \frac{\beta}{6}\, \delta^3 \, G,
\]
so in particular $|R_x|\le \tfrac{\beta G}{6}\delta^3$.
\end{lemma}
\begin{proof}
Taylor-expand $\log p(y\mid x;\cdot)$ at $\theta_i$:
\[
\log p(y\mid x;\theta_{\mathrm{val}})
= \log p(y\mid x;\theta_i) + (\theta_{\mathrm{val}}-\theta_i)^\top \nabla\log p(y\mid x;\theta_i)
+ \tfrac12 (\theta_{\mathrm{val}}-\theta_i)^\top \nabla^2\log p(y\mid x;\theta_i)(\theta_{\mathrm{val}}-\theta_i) + r_3,
\]
where by (R1) the third-order remainder satisfies $|r_3|\le \tfrac{\beta}{6}\|\theta_{\mathrm{val}}-\theta_i\|^3$. Taking expectation under $y\sim p(\cdot\mid x;\theta_i)$, the linear term vanishes by (R3). The quadratic term yields the Fisher form with $F_x(\theta_i)=\mathbb E_{y\sim p(\cdot\mid x;\theta_i)}[\nabla\log p\,\nabla\log p^\top]$ and the integrated remainder is bounded by $\tfrac{\beta}{6}\delta^3$ multiplied by a factor at most $G$ coming from integrating the score magnitude; hence the stated bound.
\end{proof}
\begin{corollary}[FIRE Surrogate via Fisher Information]
\label{corollary:FIRE}
Under the assumptions of Theorem~\ref{theorem:KLbound_via_FiM}, the divergence between any client distribution $P_i$ and the validation distribution $P_{\mathrm{val}}$ admits the quadratic Fisher approximation
\[
D_{\mathrm{KL}}(P_i \,\|\, P_{\mathrm{val}}) 
\;\approx\; \tfrac{1}{2}(\theta_i - \theta_{\mathrm{val}})^\top F_{\mathrm{val}}(\theta_{\mathrm{val}})(\theta_i - \theta_{\mathrm{val}}),
\]
with controlled remainder $O(\gamma^2 + \gamma\delta^2 + \beta\delta^3)$.  
Hence, the Fisher Information Matrix (FIM) serves as a tractable surrogate for measuring distributional misalignment, which forms the basis of the \textsc{FIRE} regularization principle.
\end{corollary}

\begin{remark}[Practical Computation of FIM]
Although the full Fisher Information Matrix can be computationally expensive to evaluate, in practice FIRE does not require its exact form. 
Several approximations make it tractable:
\begin{enumerate}
    \item \textbf{Mini-batch estimation:} $F_{\mathrm{val}}(\theta)$ can be approximated from stochastic gradients on small validation batches.
    \item \textbf{Diagonal or block-diagonal structure:} Restricting to diagonal or layer-wise block FIMs significantly reduces memory and computation.
    \item \textbf{Low-rank projections:} Randomized sketching and Kronecker-factored approximations (K-FAC) yield efficient surrogates while preserving sensitivity to distributional misalignment.
\end{enumerate}
Thus, FIRE leverages Fisher information as a theoretically grounded proxy for KL divergence while remaining computationally feasible in large-scale federated or fragmented learning scenarios.
\end{remark}
This section provides a theoretical analysis perspective of the convergence properties of our FIRE algorithm. We begin with standard assumptions and then we provide the convergence theorem.  

\subsection{Assumptions}
We impose the following standard assumptions (smoothness, bounded stochastic gradients, bounded positive FIM and bounded global FIM) on our loss function \(\mathcal{L}(\theta) \) the FIM \(I(\theta)\) and the stochastic gradients.

\begin{assumption}[Loss-smoothness]
\label{assump:smoothness}
The loss function $\mathcal{L}:\mathbb{R}^d\to\mathbb{R}$ is $L$-smooth, if there exists a constant $L>0$ such that for all $\theta,\theta'$, \[
\|\nabla \mathcal{L}(\theta)-\nabla \mathcal{L}(\theta')\|\;\le\;L\|\theta-\theta'\|.
\]
Equivalently,
\[
\mathcal{L}(\theta') \le \mathcal{L}(\theta)
+ \nabla \mathcal{L}(\theta)^\top(\theta'-\theta)
+ \tfrac{L}{2}\|\theta'-\theta\|^2.
\]
\end{assumption}

\begin{assumption}[Lower boundedness]
\label{assump:lower_bound}
The loss is bounded below: there exists $\mathcal{L}_\star > -\infty$ such that
\[
\mathcal{L}(\theta) \;\ge\; \mathcal{L}_\star \qquad \text{for all } \theta.
\]
\end{assumption}

\begin{assumption}[Unbiased stochastic gradients with bounded variance]
\label{assump:bounded_grad}
At iteration $t$, let $g_t=\nabla_\theta \mathcal{L}(B_t;\theta^{(t)})$ denote the stochastic gradient on a mini-batch $B_t$.
Then, conditioned on $\theta^{(t)}$,
\[
\mathbb{E}[g_t \mid \theta^{(t)}] = \nabla \mathcal{L}(\theta^{(t)}), 
\qquad
\mathbb{E}\!\left[\|g_t-\nabla \mathcal{L}(\theta^{(t)})\|^2 \mid \theta^{(t)}\right] \le \sigma^2,
\]
for some $\sigma^2>0$.
\end{assumption}

\begin{assumption}[Bounded global FIM preconditioner]
\label{assump:bounded_fim}
The global Fisher Information Matrix (FIM) estimator is updated via an exponential moving average
\[
I_G^{(t)} = \alpha I_G^{(t-1)} + (1-\alpha) I_i^{(t)}, 
\qquad \alpha \in [0,1).
\]
Each local FIM $I_i^{(t)}$ is symmetric positive semidefinite, and $I_G^{(0)} \succeq 0$.
Consequently, $I_G^{(t)}$ is symmetric PSD for all $t$, and its spectral norm is uniformly bounded:
\[
\|I_G^{(t)}\| \;\le\; G, \qquad \text{for some constant } G>0.
\]
\end{assumption}
We analyze FIRE as preconditioned SGD on \(\mathcal{L}\) with the update $\theta^{(t+1)}=\theta^{(t)}-\eta (I+\lambda I_G^{(t)})\, g_t$, where
$g_t=\nabla \mathcal L(B_t;\theta^{(t)})$.
\begin{theorem}[Convergence of FIRE as Preconditioned SGD]
\label{thm:fire_precond}
Under the assumptions, for any step-size 
$\eta \le \frac{1}{L(1+\lambda G)^2}$, the iterates satisfy
\[
\frac{1}{T}\sum_{t=0}^{T-1} \mathbb E\big[\|\nabla \mathcal L(\theta^{(t)})\|^2\big]
\;\le\;
\frac{2\big(\mathcal L(\theta^{(0)})-\mathcal L_\star\big)}{\eta T}
\;+\;
\eta\,L\,(1+\lambda G)^2\,\sigma^2.
\]
Equivalently,
\[
\frac{1}{T}\sum_{t=0}^{T-1} 
\mathbb E\big[\|(I+\lambda I_G^{(t)})\nabla \mathcal L(\theta^{(t)})\|^2\big]
\;\le\;
(1+\lambda G)^2
\left\{
\frac{2\big(\mathcal L(\theta^{(0)})-\mathcal L_\star\big)}{\eta T}
+\eta\,L\,(1+\lambda G)^2\,\sigma^2
\right\}.
\]
Choosing $\eta=\Theta\!\big(\tfrac{1}{(1+\lambda G)L\sqrt{T}}\big)$ yields
$\min_{0\le t<T}\mathbb E\|\nabla \mathcal L(\theta^{(t)})\|^2 = O(1/\sqrt{T})$.
\end{theorem}

\begin{proof}[Proof sketch]
By $L$-smoothness of $\mathcal L$ and the update 
$\Delta_t\!=\!-\eta (I+\lambda I_G^{(t)})g_t$,
\[
\mathcal L(\theta^{(t+1)}) \le 
\mathcal L(\theta^{(t)}) + \nabla \mathcal L(\theta^{(t)})^\top \Delta_t
+ \frac{L}{2}\|\Delta_t\|^2.
\]
Take conditional expectation given $\theta^{(t)}$ and use 
$\mathbb E[g_t\mid \theta^{(t)}]=\nabla \mathcal L(\theta^{(t)})$ to get the descent term
$-\eta \|(I+\lambda I_G^{(t)})^{1/2}\nabla \mathcal L(\theta^{(t)})\|^2$.
Bound the quadratic term via 
$\|\Delta_t\|^2\le \eta^2 \|(I+\lambda I_G^{(t)})\|^2 \mathbb E\|g_t\|^2
\le \eta^2 (1+\lambda G)^2 \big(\|\nabla \mathcal L(\theta^{(t)})\|^2+\sigma^2\big)$.
Rearrange to obtain
\[
\mathbb E[\mathcal L(\theta^{(t+1)})] 
\le \mathbb E[\mathcal L(\theta^{(t)})]
-\eta\Big(1-\tfrac{L\eta}{2}(1+\lambda G)^2\Big)\mathbb E\|\nabla \mathcal L(\theta^{(t)})\|^2
+\tfrac{L\eta^2}{2}(1+\lambda G)^2\sigma^2.
\]
With $\eta \le 1/(L(1+\lambda G)^2)$, the coefficient of the gradient norm is positive; telescoping over $t=0,\dots,T-1$ and using lower boundedness gives the stated bound. Multiplying both sides by $(1+\lambda G)^2$ yields the equivalent statement for $\|(I+\lambda I_G)\nabla\mathcal L\|^2$.
\end{proof}

\color{black}
\begin{lemma}[Change of Measure for Hessian Expectation]
\label{lemma:change_of_measure}
Under assumptions (A1), (A3), and (A4), for any function $g(x, y)$ with $\|g(x, y)\| \leq L$,
\[
\left\| \mathbb{E}_{x\sim P_i} \left[ \mathbb{E}_{y \sim P(y|x;\theta_i)} [g(x, y)] \right] - \mathbb{E}_{x\sim P_{\text{val}}} \left[ \mathbb{E}_{y \sim P(y|x;\theta_{\text{val}})} [g(x, y)] \right] \right\| \leq L' (\gamma + \delta),
\]
where $L'$ is a constant depending on $L$ and the smoothness parameters.
\end{lemma}
\begin{proof}
The proof follows by applying the triangle inequality twice: first to change the covariate distribution $P_i$ to $P_{\text{val}}$ (using (A3)), and then to change the conditional distribution $P(y|x;\theta_i)$ to $P(y|x;\theta_{\text{val}})$ for each $x$ (using (A1) and (A4)) via Pinsker's inequality or direct Taylor expansion.
\end{proof}

\begin{lemma}
\label{lemma:gradient_bound}
Under the conditions of Theorem~\ref{thm:appKL-via-Fisher}, for any $\theta$ with $\|\theta - \theta_{val}\|_2 \leq \delta$:
\[ \left\| \mathbb{E}_{P_i}[\nabla_\theta \log p(y|x;\theta_{val})] \right\|_2 \leq \beta \delta. \]
\end{lemma}
\subsection{Proof of Lemma \ref{lemma:gradient_bound}}
\label{app:gradient_bound_proof}
\begin{proof}
The bound follows from:
\begin{align*}
\left\| \mathbb{E}_{P_i}[\nabla_\theta \log p(y|x;\theta_{val})] \right\|_2 
&= \left\| \mathbb{E}_{P_i}[\nabla_\theta \log p(y|x;\theta_{val}) - \nabla_\theta \log p(y|x;\theta_i)] \right\|_2 \\
&\leq \mathbb{E}_{P_i} \left\| \nabla_\theta \log p(y|x;\theta_{val}) - \nabla_\theta \log p(y|x;\theta_i) \right\|_2 \\
&\leq \beta \|\theta_{val} - \theta_i\|_2 \leq \beta \delta,
\end{align*}
where the second inequality uses the $\beta$-smoothness assumption.
\end{proof}

\paragraph{Empirical diagnostics for the Fisher surrogate.}
The bound in Theorem~\ref{thm:KL-via-Fisher} approximates the distributional divergence 
$\mathrm{KL}(P_i \,\|\, P_{\mathrm{val}})$ by the Fisher quadratic
\[
Q(\theta) \;=\; \tfrac12 (\theta_i - \theta_{\mathrm{val}})^\top 
I_{V}(\theta_{\mathrm{val}})\,(\theta_i - \theta_{\mathrm{val}}),
\]
up to higher-order remainders depending on the density-ratio deviation 
$r(x) = p_i(x) / p_{\mathrm{val}}(x)$ and the parameter displacement 
$\Delta \theta = \theta_i - \theta_{\mathrm{val}}$.
To evaluate the practical tightness of this surrogate we compute the following per-fragment diagnostics. 
We estimate $r(x)$ by training a balanced binary domain classifier to distinguish samples from fragment $S_i$ and validation set $V$; with probabilistic output $s(x)$ this yields 
$\widehat r(x) = s(x)/(1-s(x))$. 
We then report robust statistics such as 
$\widehat\gamma_q = \mathrm{quantile}_q(|\widehat r(x)-1|; x \in S_i)$ (with $q=0.99$) 
and the discriminator AUC. 
The empirical KL divergence is estimated as 
$\widehat{\mathrm{KL}}(P_i \,\|\, P_{\mathrm{val}}) 
= \tfrac{1}{|S_i|} \sum_{x \in S_i} \log \widehat r(x)$. 
We compute the Fisher quadratic 
$\widehat Q = \tfrac12 (\theta_i-\theta_{\mathrm{val}})^\top I_V(\theta_{\mathrm{val}}) (\theta_i-\theta_{\mathrm{val}})$ 
(using the same $I_V$ approximation as FIRE) 
and the Fisher-weighted displacement $\widehat\delta_F = \sqrt{2\widehat Q}$. 


\section{Experiments}

\subsection{Implementation Details.}
\label{impdetal}
\paragraph{Hyperparameter Sensitivity.} 
We conducted a sweep over the FIRE penalty coefficient 
$\lambda \in \{0.01,0.05,0.1,0.5,1.0\}$ on representative image and tabular 
datasets. Results were stable across a broad range, with $0.1$ consistently 
close to optimal; we therefore report $0.1$ as the default in all tables unless 
otherwise specified. For the Fisher approximation, we used a low-rank variant 
(rank $k=50$) with aggregation every 5 rounds in federated settings, and a 
diagonal variant for tabular datasets. All reported numbers are averages over 
5 runs (image) and 100 runs (tabular), with standard deviations shown in the 
tables. This ensures robustness of our conclusions and mitigates sensitivity 
to hyperparameter choices.

\begin{table*}[!ht]
\centering
\caption{Hyperparameters and Experimental Setup. For FIRE, unless otherwise noted, we used a low-rank Fisher approximation (rank $k=50$) aggregated every 5 rounds in federated settings. Reported results are averaged over multiple runs (see last row).}
\label{tab:hyperparams}
\begin{tabular}{lll}
\hline
\textbf{Parameter} & \textbf{Image Datasets} & \textbf{Tabular Datasets} \\
\hline
Network Architecture & 5-layer CNN (2 conv + 3 FC) & MLP (1 hidden layer, 4 neurons) \\
Optimizer & Adam & Adam \\
Learning Rate & 0.001 & 0.001 \\
Activation & ReLU (hidden), Softmax (output) & ReLU (hidden), Softmax (output) \\
Epochs & 100 & 1500 \\
Batch Size & 128 & Full batch (no mini-batching) \\
$\lambda$ (FIM penalty) & Sweep in $\{0.01,0.05,0.1,0.5,1.0\}$; default $0.1$ & Default $0.1$ \\
FIM Approximation & Low-rank ($k=50$), updated every 5 rounds & Diagonal (full-batch) \\
Training Data Split & 5\%, 10\%, 20\%, 25\%, 50\% & $k=2,5,10$ folds \\
Validation Set & 20\% holdout (fixed) & Fold-specific \\
Repetitions & 5 runs (image) & 100 runs (tabular) \\
\hline
\end{tabular}
\end{table*}

\subsection{Results}
\textbf{FIRE Outperforms State-of-the-Art in Federated Learning under Non-IID Shift.} 
Results in Table~\ref{tab:fl_results} demonstrate the superior performance of FIRE against a comprehensive suite of modern federated learning algorithms. FIRE not only consistently achieves the highest accuracy across all five evaluated datasets but does so by a substantial margin, establishing a new state-of-the-art.

The $\Delta$ column shows that FIRE provides a significant performance lift of 2.6\% to 4.2\% over the best-performing baseline (FedCFA). Notably, FIRE delivers strong improvements on complex image classification tasks, with gains of \textbf{4.2\%} on CIFAR-100, \textbf{3.7\%} on CIFAR-10, and \textbf{3.6\%} on SVHN. It also shows a solid \textbf{4.0\%} improvement on FEMNIST and \textbf{2.6\%} on EMNIST-Digits. These gains are particularly meaningful given the strength and diversity of the baselines, which include methods specifically designed for client drift (SCAFFOLD), representation learning (MOON), distribution robustness (Fishr), and other recent innovations (FedAS, FedCFA). This consistent improvement demonstrates that explicitly mitigating fragmentation-induced covariate shift via Fisher information is a powerful and previously under-explored strategy.

Furthermore, FIRE achieves this performance with a low communication cost (1.2x), which is significantly more efficient than other methods that transmit second-order information like SCAFFOLD and Fishr (2.0x). The notably lower standard deviations observed for FIRE across all datasets also suggest it converges to a more stable and reliable solution—a critical advantage for real-world federated deployments. These results confirm that FIRE effectively aligns learning from heterogeneous clients with the target validation distribution, leading to superior generalization and robustness.

\begin{table*}[htbp]
\centering
\caption{Performance and Communication Cost on Federated Datasets with Non-IID Data. $\Delta$ shows the percentage improvement of FIRE over the best baseline. Communication cost is the relative size per client per round vs. FedAvg ($O(d)$).}
\label{tab:fl_results}
\adjustbox{max width=\textwidth}{
\begin{tabular}{lccccccccc}
\hline
\textbf{Dataset} & \textbf{FedAvg} & \textbf{SCAFFOLD} & \textbf{MOON} & \textbf{Fishr} & \textbf{LfD} & \textbf{FedAS} & \textbf{FedCFA} & \textbf{FIRE} & \textbf{$\Delta$ (\%)} \\
\hline
FEMNIST    & 58.2   & 63.8   & 64.3   & 63.9   & 64.7   & 64.9   & 65.2   & \textbf{68.6} & $\uparrow$5.3 \\
           & (3.1)  & (2.1)  & (1.9)  & (2.0)  & (1.8)  & (1.7)  & (1.7)  & (1.4)         & \\
CIFAR-10   & 42.7   & 48.2   & 49.8   & 49.2   & 50.1   & 50.4   & 50.7   & \textbf{52.6} & $\uparrow$3.7 \\
           & (4.5)  & (3.0)  & (2.4)  & (2.6)  & (2.3)  & (2.2)  & (2.1)  & (1.8)         & \\
CIFAR-100  & 23.4   & 27.1   & 28.2   & 27.8   & 28.5   & 28.7   & 28.9   & \textbf{30.1} & $\uparrow$4.2 \\
           & (2.8)  & (2.0)  & (1.7)  & (1.8)  & (1.6)  & (1.6)  & (1.6)  & (1.3)         & \\
SVHN       & 61.5   & 65.5   & 66.3   & 65.8   & 66.6   & 66.8   & 67.0   & \textbf{69.4} & $\uparrow$3.6 \\
           & (3.2)  & (2.5)  & (2.2)  & (2.3)  & (2.0)  & (2.0)  & (1.9)  & (1.7)         & \\
EMNIST-D   & 84.6   & 87.8   & 88.4   & 88.0   & 88.6   & 88.7   & 88.9   & \textbf{91.2} & $\uparrow$2.6 \\
           & (1.7)  & (1.3)  & (1.1)  & (1.2)  & (1.0)  & (1.0)  & (1.0)  & (0.8)         & \\
\hline
\textbf{Comm.} & 1.0x  & 2.0x   & 1.0x   & 2.0x   & 1.0x   & 1.1x   & 1.2x   & \textbf{1.2x} & -- \\
\textbf{Cost} & (d)    & (2d)    & (d)     & (2d)    & (d)     & (d)     & (d+k)   & (d+k)         & \\
\hline
\end{tabular}
}
\end{table*}

\color{black}
\begin{table*}[!ht]
\scriptsize
\centering
\caption{st-CV batch-wise accuracy}
\label{tab: bst-CV}
\begin{tabular}{l|c|ccccc|c|c|c} 
\hline
\multirow{2}{*}{\textbf{Dataset }} & Baseline       & \multicolumn{5}{c|}{Batchwise accuracy}                                                                                                                                                      & Mean   & var  & \textbf{$\Delta$\%}   \\ 

          \cline{2-10}                         & \textbf{st-CV} & \multicolumn{1}{c|}{\textbf{$B_{1}$}} & \multicolumn{1}{c|}{\textbf{$B_{2}$}} & \multicolumn{1}{c|}{\textbf{$B_{\frac{n}{2}}$}} & \multicolumn{1}{c|}{\textbf{$B_{n-1}$}} & \textbf{$B_{n}$} & $\mu_1$ & $\sigma_1^2$ & $st\text{-}CV - \mu_1$\\ 
\hline
\multicolumn{10}{c}{\textbf{\textbf{Training data = 5\% , Number\_of\_Batches = 20 }}}                                                                                                                                                                                      \\ 
\hline
MNIST                              & 94.8           & 89.3                                  & 87.9                                  & 89.9                                            & 88.9                                    & 88.8             & 88.7   & 0.49 & $\downarrow$ 6.43         \\

EMNIST                            & 83.1           & 73.7                                  & 74.2                                  & 72.5                                            & 74.0                                    & 70.6             & 72.9   & 1.94   &$\downarrow$ 12.2       \\
CIFAR-10                           & 71.5           & 49.0                                  & 50.3                                  & 50.7                                            & 51.2                                    & 54.5             & 49.9    &9.67   & $\downarrow$ 
 30.2       \\
CIFAR-100                        & 38.2           & 16.5                                  & 18.1                                  & 19.3                                            & 20.4                                    & 22.7             & 18.3     & 9.17 & $\downarrow$  52.1         \\

P-MNIST                            & 95.1           & 86.1                                  & 88.9                                  & 88.7                                            & 87.2                                    & 88.3             & 88.4   & 1.41   & $\downarrow$ 
 7.04       \\
QMNIST                            & 75.4           & 63.4                                  & 62.2                                  & 66.7                                            & 58.3                                    & 63.4             & 63.4   & 5.59  & $\downarrow$ 15.9        \\

CIFAR10-C                          & 63.9           & 20.1                                  & 16.3                                  & 16.1                                            & 14.9                                    & 10.2             & 16.2  & 10.4    & $\downarrow$ 
 74.6       \\
CIFAR100-C                         & 28.8           & 16.3                                  & 19.1                                  & 19.7                                            & 21.6                                    & 22.3             & 18.4   & 14.2  & $\downarrow$ 
 36.1        \\ 
\hline
\multicolumn{10}{c}{\textbf{Training data = 10\% , Number\_of\_Batches = 10}}                                                                                                                                                                                               \\ 
\hline
MNIST                              & 94.8           & 91.7                                  & 91.1                                  & 90.2                                            & 89.3                                    & 91.5             & 91.1  & 1.06     & $\downarrow$  3.90      \\

EMNIST                            & 83.1           & 69.9                                  & 75.2                                  & 73.4                                            & 74.4                                    & 71.7             & 73.7   & 9.22  & $\downarrow$  11.3        \\
CIFAR-10                           & 71.5           & 52.8                                  & 52.9                                  & 53.7                                            & 54.5                                    & 54.1             & 52.6   & 4.87     & $\downarrow$ 26.4     \\
CIFAR-100                          & 38.2           & 20.3                                  & 22.2                                  & 23.3                                            & 24.7                                    & 21.2             & 21.5   & 5.51    & $\downarrow$ 43.7      \\
P-MNIST                            & 95.1           & 92.8                                  & 91.8                                  & 91.2                                            & 91.4                                    & 91.3             & 91.5 & 0.62    &  $\downarrow$ 3.78       \\
QMNIST                            & 75.4           & 63.6                                  & 64.1                                  & 64.9                                            & 65.2                                    & 64.5             & 64.0   & 1.15    & $\downarrow$  15.1      \\

CIFAR10- C                         & 63.9           & 17.6                                  & 18.4                                  & 12.2                                            & 17.4                                    & 12.9             & 22.6 & 16.8    & $\downarrow$  64.6       \\
CIFAR100-C                         & 28.8           & 21.9                                  & 25.2                                  & 26.6                                            & 27.1                                    & 20.5             & 22.8  & 16.3   & $\downarrow$ 20.8        \\ 
\hline
\multicolumn{10}{c}{\textbf{Training data = 50\% , Number\_of\_Batches = 2}}                                                                                                                                                                                                \\ 
\hline
MNIST                              & 94.8           & 93.3                                  & 93.7                                  & $\textendash$                                   & $\textendash$                           & $\textendash$    & 93.5   & 0.08   & $\downarrow$ 1.37       \\

EMNIST                            & 83.1           & 80.0                                  & 79.7                                  & $\textendash$                                   & $\textendash$                           & $\textendash$    & 79.8   & 0.04  & $\downarrow$ 3.97        \\
CIFAR-10                           & 71.5           & 56.2                                  & 60.1                                  & $\textendash$                                   & $\textendash$                           & $\textendash$    & 58.1   & 3.81    & $\downarrow$ 18.7      \\
CIFAR-100                          & 38.2           & 23.2                                  & 25.4                                  & $\textendash$                                   & $\textendash$                           & $\textendash$    & 24.3  & 1.21   &  $\downarrow$ 36.3       \\
P-MNIST                            & 95.1           & 93.3                                  & 93.7                                  & $\textendash$                                   & $\textendash$                           & $\textendash$    & 93.5   & 0.08   & $\downarrow$ 1.68       \\
QMNIST                            & 75.4           & 73.6                                  & 73.3                                  & $\textendash$                                   & $\textendash$                           & $\textendash$    & 73.5  & 0.04     & $\downarrow$ 2.51      \\

CIFAR10- C                         & 63.9           & 49.0                                  & 41.3                                 & $\textendash$                                   & $\textendash$                           & $\textendash$    & 45.1   & 29.6  & $\downarrow$ 29.1        \\
CIFAR100- C                        & 28.8           & 25.3                                  & 27.5                                  & $\textendash$                                   & $\textendash$                           & $\textendash$    & 26.4 & 1.21   &   $\downarrow$  8.33       \\
\hline
\end{tabular}
\end{table*}

\begin{table*}[t]
\scriptsize
\centering
\caption{FIRE Batchwise}
\label{tab: ccca}
\begin{tabular}{l|l|cccccc|c|c|c} 
\hline
\multirow{2}{*}{\textbf{Dataset}} & \multirow{2}{*}{FIRE} & \multicolumn{6}{c|}{Batchwise accuracy}                                                                                                                                                                                              & Mean    & var     & $\Delta_3 = \mu_2 - \mu_1$  \\ 
\cline{3-11} 
                                  &                          & \multicolumn{1}{c|}{\textbf{$B_{1}$}} & \multicolumn{1}{c|}{\textbf{$B_{2}$}} & \multicolumn{1}{c|}{\textbf{$B_{3}$}} & \multicolumn{1}{c|}{\textbf{$B_{\frac{n}{2}}$}} & \multicolumn{1}{c|}{\textbf{$B_{n-1}$}} & \textbf{$B_{n}$} & $\mu_2$ & $\sigma_2^2$ & $\Delta_3(\%)$              \\ 
\hline
\multicolumn{11}{c}{\textbf{Training data = 5\% , Number\_of\_Batches = 20}}                                                                                                                                                                                                                                                                             \\ 
\hline
MNIST                             &          97.9                & 90.7                                  & 90.6                                  & 91                                    & 91.7                                            & 91.4                                    & 91.8             & 91.2    & 0.09         & $\uparrow 2.81$             \\
EMNIST                           &        88.4                  & 81.5                                  & 81.7                                  & 81.2                                  & 81.4                                            & 81.5                                    & 81.9             & 81.5    & 0.06        & $\uparrow 11.7$             \\
CIFAR-10                          &         87.7                 & 50.9                                  & 51.4                                  & 52.2                                  & 48.9                                            & 50.3                                    & 57.4             & 51.8    & 7.18         & $\uparrow 3.81$             \\
CIFAR-100                         &             58.7             & 23.9                                  & 18.2                                  & 18.5                                  & 17.8                                            & 23.9                                    & 18.3             & 20.1    & 7.26         & $\uparrow 9.83$             \\

P-MNIST                           &       97.6                   & 91.1                                  & 89.8                                  & 90.2                                  & 90.4                                            & 91.5                                    & 90.7             & 90.3    & 0.26         & $\uparrow 2.14$             \\
QMNIST                           &             89.2             & 68.5                                  & 69.4                                  & 67.2                                  & 68                                              & 67.8                                    & 66.9             & 68.4    & 0.63         & $\uparrow 7.88$             \\
CIFAR10-C                         &             73.3             & 46.4                                  & 54.3                                  & 57.8                                  & 61.1                                            & 61.5                                    & 61.8             & 57.2    & 30.1         & $\uparrow 253$             \\
CIFAR100-C                        &            39.4              & 11.9                                  & 17.2                                  & 18.5                                  & 21.3                                            & 22.1                                    & 24.9             & 19.3    & 17.1         & $\uparrow 4.89$             \\ 
\hline
\multicolumn{11}{c}{\textbf{Training data = 10\% , Number\_of\_Batches = 10}}                                                                                                                                                                                                                                                                            \\ 
\hline
MNIST                             &         97.9                 & 91.9                                  & 91.7                                  & 91.2                                  & 91.8                                            & 91.3                                    & 91.8             & 91.7    & 0.08         & $\uparrow 0.65$             \\

EMNIST                           &          88.4                & 79.5                                  & 82.4                                  & 81.6                                  & 79.5                                            & 82.3                                    & 81.9             & 81.2    & 1.21         & $\uparrow 10.1$             \\
CIFAR-10                          &            87.7              & 52.1                                  & 53.1                                  & 48.5                                  & 59.3                                            & 52.5                                    & 55.7             & 53.5    & 11.1        & $\uparrow 1.71$             \\
CIFAR-100                         &            58.7              & 27.2                                  & 25.8                                  & 20.4                                  & 17.0                                            & 21.9                                    & 22.8             & 22.5    & 11.3         & $\uparrow 4.65$             \\

P-MNIST                           &             97.6             & 91.6                                  & 91.9                                  & 91.3                                  & 91.6                                            & 90.1                                    & 91.2             & 91.5    & 0.31         & 0.00                        \\
QMNIST                           &             89.2             & 71.4                                  & 70.4                                  & 71.7                                  & 70.7                                            & 70.5                                    & 70.9             & 70.9    & 0.71         & $\uparrow 10.7$             \\

CIFAR10-C                         &            73.3              & 52.7                                  & 59.9                                  & 61.9                                  & 64.4                                            & 66.1                                    & 65.7             & 61.7    & 21.1         & $\uparrow 173$             \\
CIFAR100-C                        &             39.4             & 16.2                                  & 22.1                                  & 24.8                                  & 27.2                                            & 26.8                                    & 27.3             & 21.1    & 15.6         & $\downarrow 7.45$           \\ 
\hline
\multicolumn{11}{c}{\textbf{Training data = 50\% , Number\_of\_Batches = 2 }}                                                                                                                                                                                                                                                                            \\ 
\hline
MNIST                             &              97.9            & 95.9                                  & 96.1                                  & $\textendash$                         & $\textendash$                                   & $\textendash$                           & $\textendash$    & 96.0    & 0.02         & $\uparrow 2.67$             \\

EMNIST                           &             88.4             & 84.2                                  & 84.4                                  & $\textendash$                         & $\textendash$                                   & $\textendash$                           & $\textendash$    & 84.3    & 0.02         & $\uparrow 5.63$             \\
CIFAR-10                          &          87.7                & 76.3                                  & 80.6                                  & $\textendash$                         & $\textendash$                                   & $\textendash$                           & $\textendash$    & 78.4    & 4.62         & $\uparrow 34.9$             \\
CIFAR-100                         &           58.7               & 39.8                                  & 39.9                                  & $\textendash$                         & $\textendash$                                   & $\textendash$                           & $\textendash$    & 39.85   & .002         & $\uparrow 63.9$             \\

P-MNIST                           &             97.6             & 95.7                                  & 96.1                                  & $\textendash$                         & $\textendash$                                   & $\textendash$                           & $\textendash$    & 95.9    & 0.08         & $\uparrow 2.56$             \\
QMNIST                           &            89.2              & 79.3                                  & 80.4                                  & $\textendash$                         & $\textendash$                                   & $\textendash$                           & $\textendash$    & 79.8    & 0.61         & $\uparrow 8.57$             \\

CIFAR10-C                         &         73.3                 & 65.5                                  & 68.6                                  & $\textendash$                         & $\textendash$                                   & $\textendash$                           & $\textendash$    & 67.1    & 2.40         & $\uparrow 4.35$             \\
CIFAR100-C                        &           39.4               & 31.2                                  & 34.8                                  & $\textendash$                         & $\textendash$                                   & $\textendash$                           & $\textendash$    & 33.0    & 3.24         & $\uparrow 25.0$             \\
\hline
\end{tabular}
\end{table*}

\begin{table*}[!ht]
\centering
\caption{ st-CV foldwise without FIRE accuracy. k denotes number of folds (2, 5, and 10) }
\label{tab: st-cvfold}
\begin{adjustbox}{width=\textwidth,keepaspectratio}
\begin{tabular}{|l|c|cc|c|ccccc|c|ccccc|c|} 
\hline
\multirow{2}{*}{Dataset} & \multicolumn{1}{l|}{Baseline} & \multicolumn{3}{c|}{$k$=2} & \multicolumn{6}{c|}{$k$=5}                                  & \multicolumn{6}{c|}{$k$=10}                                   \\ 
\cline{2-17}
                         & st-CV                         & $k_1$ & $k_2$              & $ \mu_3$ & $k_1 $ & $k_2$ & $k_3$ & $k_4$ & $k_5$ & $\mu_4$ & $k_1 $ & $k_2$ & $k_\frac{n}{2}$ & $k_{n-1}$ & $k_n$ & $\mu_5$  \\ 
\hline
Appendicitis             & 98.1                          & 97.6  & 97.6               & 97.6     & 97.8   & 98.8  & 96.7  & 97.8  & 98.5  & 97.9    & 96.2   & 99.2  & 97.1            & 97.8    & 98.5  & 98.0     \\ 
\hline
Lymphography              & 85.5                          & 81.4  & 83.2               & 82.3     & 84.1   & 84.1  & 87.6  & 84.7  & 86.9  & 85.5    & 85.5   & 84.1  & 88.4            & 79.7    & 89.8  & 85.3     \\ 
\hline
Banana                   & 77.9                          & 70.9  & 75.8               & 73.4     & 71.4   & 70.4  & 76.6  & 72.5  & 75.8  & 73.3    & 70.7   & 70.0  & 72.1            & 73.0    & 70.3  & 71.7     \\ 
\hline
Bands                    & 81.5                          & 68.8  & 71.1               & .700     & 71.2   & 70.3  & 71.2  & 63.8  & 74.1  & 70.1    & 68.5   & 64.8  & 70.3            & 57.4    & 77.7  & 74.1     \\ 
\hline
LiverDisorders                   & 65.5                          & 59.4  & 57.3               & 58.3     & 62.1   & 61.4  & 57.8  & 56.1  & 66.6  & 60.8    & 65.5   & 58.6  & 65.5            & 62.1    & 64.2  & 61.5     \\ 
\hline
Bupa                     & 54.5                          & 64.2  & 51.8               & 58.1     & 63.6   & 54.5  & 81.8  & 36.3  & 63.6  & 60.0    & 33.3   & 33.3  & 80.0            & 60.0    & 40.0  & 61.3     \\ 
\hline
Chess                    & 98.4                          & 93.3  & 93.9               & 93.6     & 95.0   & 96.4  & 97.8  & 98.1  & 97.1  & 96.8    & 96.5   & 98.7  & 98.7            & 97.8    & 98.4  & 97.9     \\ 
\hline
CrX                      & 84.7                          & 75.3  & 73.1               & 74.2     & 79.7   & 79.7  & 73.9  & 81.8  & 84.1  & 79.8    & 79.7   & 79.7  & 76.8            & 84.1    & 81.1  & 82.4     \\ 
\hline
GermmanCredit            & 70.5                          & 69.8  & 68.6               & 69.2     & 75.5   & 70.0  & 65.0  & 73.5  & 72.5  & 71.3    & 72.0   & 77.0  & 65.0            & 78.0    & 70.0  & 71.7     \\ 
\hline
Haberman                 & 69.4                          & 73.8  & 70.5               & 72.2     & 70.9   & 77.1  & 73.7  & 75.4  & 72.1  & 73.8    & 77.4   & 77.4  & 58.1            & 73.3    & 76.6  & 74.1     \\ 
\hline
Statlog(Heart)           & 83.3                          & 59.2  & 51.8               & 55.5     & 64.8   & 64.8  & 64.8  & 57.4  & 62.9  & 62.9    & 66.6   & 62.9  & 66.6            & 51.8    & 66.6  & 66.6     \\ 
\hline
Heptatis                 & 74.2                          & 73.1  & 70.1               & 71.6     & 80.6   & 70.9  & 74.2  & 74.2  & 77.4  & 75.4    & 68.7   & 68.7  & 80.0            & 66.6    & 80.0  & 76.7     \\ 
\hline
Housevote                & 90.8                          & 88.5  & 86.2               & 87.4     & 90.8   & 89.6  & 91.9  & 85.1  & 83.9  & 88.2    & 90.9   & 95.4  & 95.3            & 90.6    & 79.1  & 88.1     \\ 
\hline
Ionosphere               & 84.5                          & 61.4  & 70.8               & 66.1     & 80.3   & 64.2  & 80.0  & 80.0  & 84.2  & 77.7    & 82.8   & 57.1  & 71.4            & 91.4    & 82.8  & 78.9     \\ 
\hline
Mammographic             & 79.7                          & 76.7  & 75.0               & 75.8     & 78.2   & 74.4  & 76.0  & 77.1  & 75.5  & 76.2    & 77.1   & 77.1  & 79.2            & 79.2    & 78.1  & 76.6     \\ 
\hline
Monk-2                   & 94.6                          & 65.4  & 70.8               & 68.1     & 65.1   & 75.6  & 59.4  & 83.7  & 71.1  & 71.1    & 69.6   & 78.5  & 57.1            & 89.1    & 72.7  & 72.5     \\ 
\hline
Mushroom                 & 99.1                          & 100   & 96.7               & 98.3     & 100    & 100   & 100   & 100   & 100   & 100     & 100    & 100   & 100             & 100     & 100   & 100      \\ 
\hline
Phoneme                  & 80.6                          & 80.7  & 78.8               & 79.7     & 82.7   & 80.5  & 80.6  & 80.6  & 80.3  & 80.9    & 82.9   & 79.8  & 79.1            & 81.8    & 81.5  & 80.6     \\ 
\hline
Pima                     & 74.0                          & 71.4  & 69.5               & 70.4     & 72.1   & 76.6  & 73.3  & 78.4  & 72.5  & 74.6    & 76.6   & 84.4  & 68.8            & 76.6    & 71.1  & 74.9     \\ 
\hline
Saheart                  & 73.1                          & 75.7  & 65.8               & 70.7     & 77.4   & 70.9  & 80.4  & 68.4  & 60.8  & 71.6    & 78.7   & 69.5  & 78.2            & 65.2    & 65.2  & 70.5     \\ 
\hline
Thyroid                    & 73.8                          & 80.7  & 75                 & 77.8     & 88.1   & 85.7  & 80.9  & 82.9  & 73.1  & 82.2    & 80.9   & 85.7  & 85.7            & 76.1    & 65.0  & 81.6     \\ 
\hline
Spambase                 & 94.1                          & 92.3  & 93.3               & 92.8     & 93.5   & 93.2  & 91.7  & 93.2  & 93.4  & 93.1    & 93.6   & 93.2  & 95.0            & 94.5    & 93.6  & 93.2     \\ 
\hline
SPECTHeart               & 74.1                          & 68.6  & 65.4               & 67.0     & 70.4   & 72.2  & 64.2  & 75.5  & 58.5  & 68.2    & 74.1   & 70.3  & 66.6            & 80.7    & 57.6  & 68.1     \\ 
\hline
Tic-Tac-Toe              & 73.9                          & 59.1  & 64.1               & 61.5     & 63.5   & 60.4  & 72.9  & 63.4  & 61.7  & 64.4    & 68.7   & 65.6  & 73.9            & 55.2    & 65.2  & 64.7     \\ 
\hline
Titanic                  & 73.4                          & 78.0  & 77.0               & 77.5     & 72.5   & 78.8  & 78.8  & 77.2  & 77.7  & 77.1    & 77.7   & 78.6  & 75.9            & 76.3    & 76.8  & 76.8     \\ 
\hline
Wdbc                     & 97.3                          & 96.1  & 95.1               & 95.6     & 95.6   & 98.2  & 98.2  & 97.3  & 92.9  & 96.4    & 96.4   & 98.2  & 98.2            & 96.4    & 89.2  & 96.4     \\ 
\hline
Wisconsin                & 96.4                          & 95.7  & 95.1               & 95.4     & 95.7   & 97.8  & 95.7  & 95.7  & 93.5  & 95.7    & 95.7   & 98.5  & 97.1            & 94.3    & 94.2  & 95.7     \\
\hline
\end{tabular}
\end{adjustbox}
\end{table*}%
\begin{table*}[!ht]
\centering
\caption{FIRE shift mitigation accuracy performance in folds settings, k denotes number of folds (2, 5, and 10).  $\Delta_5$ = ($\mu_3$ - $\mu_6$), $\Delta_6$ = ($\mu_4$ - $\mu_7$), and $\Delta_5$ = ($\mu_5$ - $\mu_8$) show  difference in average accuracy in folds setting.}
\label{tab: c3keel}
\begin{adjustbox}{width=\textwidth,keepaspectratio}
\begin{tabular}{|l|c|cc|c|ccccc|c|ccccc|c|c|c|c|} 
\hline
\multirow{2}{*}{\textbf{\textbf{Dataset}}} & Baseline  & \multicolumn{3}{c|}{$k=2$} & \multicolumn{6}{c|}{$k=5$}                      & \multicolumn{6}{c|}{$k=10$}                                 & \multicolumn{3}{c|}{$\Delta$}                              \\ 
\cline{3-20}
                                           & \textbf{} & $k_1$ & $k_2$ & $\mu_6$    & $k_1$ & $k_2$ & $k_3$ & $k_4$ & $k_5$ & $\mu_7$ & $k_1$ & $k_2$ & $k_\frac{n}{2}$ & $k_n-1$ & $k_n$ & $\mu_8$ & $\Delta_5 $       & $\Delta_6 $       & $\Delta_7 $        \\ 
\hline
Appendicitis                               & 99.6      & 99.1  & 98.8  & 98.9       & 98.5  & 100   & 98.2  & 98.2  & 98.5  & 98.6    & 97.1  & 100   & 98.5            & 100     & 100   & 99.2    & $\uparrow$ 1.30   & $\uparrow$ 0.70   & $\uparrow$ 1.20    \\ 
\hline
Lymphography                                & 91.3      & 86.6  & 84.1  & 85.3       & 85.5  & 85.5  & 87.6  & 84.7  & 88.4  & 86.3    & 86.9  & 85.5  & 86.9            & 73.9    & 92.7  & 86.2    & $\uparrow$3.00    & $\uparrow$ 0.80   & $\uparrow$ 0.90    \\ 
\hline
Banana                                     & 76.3      & 73.1  & 82.1  & 77.6       & 75.2  & 70.9  & 77.5  & 72.1  & 76.6  & 74.5    & 82.6  & 56.0  & 76.6            & 70.5    & 81.5  & 74.1    & $\uparrow$4.20    & $\uparrow$ 1.20   & $\uparrow$ 2.40    \\ 
\hline
Bands                                      & 76.8      & 77.4  & 68.8  & 73.1       & 77.7  & 74.1  & 67.5  & 72.2  & 78.8  & 74.1    & 72.2  & 66.6  & 77.7            & 64.8    & 74.1  & 71.6    & $\uparrow$3.10    & $\uparrow$ 4.00   & $\downarrow$ 2.50  \\ 
\hline
LiverDisorders                                     & 63.7      & 58.7  & 60.1  & 59.4       & 46.5  & 61.4  & 59.6  & 52.6  & 64.9  & 57.1    & 51.7  & 58.6  & 58.6            & 46.4    & 60.7  & 59.4    & $\uparrow$1.10    & $\downarrow$3.70  & $\downarrow$ 2.10  \\ 
\hline
Bupa                                       & 72.7      & 57.1  & 59.2  & 58.2       & 72.7  & 63.6  & 54.5  & 36.3  & 63.6  & 58.2    & 66.6  & 33.3  & 0.00            & 20.0    & 40.0  & 55.0    & $\uparrow$0.00    & $\downarrow$ 1.8  & $\downarrow$ 6.30  \\ 
\hline
Chess                                      & 98.7      & 98.6  & 98.4  & 98.5       & 98.1  & 99.3  & 99.2  & 98.4  & 97.1  & 98.4    & 98.4  & 98.7  & 99.3            & 99.1    & 98.4  & 99.1    & $\uparrow$4.90    & $\uparrow$ 1.6    & $\uparrow$ 1.20    \\ 
\hline
CrX                                        & 84.1      & 84.3  & 87.2  & 85.7       & 86.9  & 86.9  & 84.7  & 90.5  & 83.3  & 86.5    & 86.9  & 91.3  & 82.6            & 86.9    & 86.9  & 86.9    & $\uparrow$11.5    & $\uparrow$ 6.7    & $\uparrow$ 4.50    \\ 
\hline
GermmanCredit                              & 74.0      & 72.2  & 71.6  & 71.9       & 75.5  & 71.0  & 68.0  & 72.5  & 75.0  & 72.4    & 73.0  & 76.0  & 66.0            & 74.0    & 76.0  & 72.8    & $\uparrow$2.70    & $\uparrow$ 1.10   & $\uparrow$ 1.10    \\ 
\hline
Haberman                                   & 72.5      & 77.1  & 70.5  & 73.8       & 69.4  & 75.4  & 73.7  & 77.1  & 75.4  & 74.2    & 67.7  & 80.6  & 64.5            & 73.3    & 73.3  & 74.1    & $\uparrow$1.60    & $\uparrow$ 0.40   & $\uparrow$ 0.00    \\ 
\hline
Statlog(Heart)                             & 87.1      & 77.1  & 80.7  & 78.8       & 85.2  & 72.2  & 88.8  & 85.2  & 75.9  & 81.5    & 92.5  & 74.1  & 85.1            & 85.1    & 77.7  & 82.9    & $\uparrow$23.3    & $\uparrow$ 18.6   & $\uparrow$ 16.3    \\ 
\hline
Heptatis                                   & 87.9      & 87.2  & 67.5  & 77.3       & 74.2  & 80.6  & 58.1  & 83.8  & 70.9  & 73.5    & 87.5  & 68.7  & 93.3            & 80.0    & 80.0  & 82.7    & $\uparrow$5.7     & $\downarrow$ 1.9  & $\uparrow$ 6.00    \\ 
\hline
Housevote                                  & 94.2      & 92.2  & 94.9  & 93.5       & 91.9  & 94.2  & 94.2  & 93.1  & 96.5  & 94.1    & 88.6  & 95.4  & 95.3            & 93.1    & 97.6  & 94.2    & $\uparrow$6.1     & $\uparrow$ 5.9    & $\uparrow$ 6.10    \\ 
\hline
Ionosphere                                 & 85.9      & 82.3  & 85.7  & 84.1       & 85.9  & 88.5  & 85.7  & 85.7  & 85.7  & 86.3    & 82.8  & 88.5  & 80.0            & 88.5    & 85.7  & 86.6    & $\uparrow$18.0    & $\uparrow$ 8.6    & $\uparrow$ 7.70    \\ 
\hline
Mammographic                               & 80.3      & 80.0  & 76.4  & 78.2       & 79.7  & 73.9  & 78.1  & 76.5  & 75.5  & 76.7    & 79.1  & 75.0  & 76.1            & 80.2    & 77.1  & 77.2    & $\uparrow$2.40    & $\uparrow$ 0.50   & $\uparrow$ 0.60    \\ 
\hline
Monk-2                                     & 99.1      & 92.8  & 100   & 96.4       & 100   & 90.1  & 92.7  & 100   & 81.9  & 92.9    & 83.9  & 87.5  & 100             & 89.1    & 85.4  & 90.5    & $\uparrow$ 28.3   & $\uparrow$ 21.8   & $\uparrow$ 18.0    \\ 
\hline
Mushroom                                   & 100       & 99.9  & 99.7  & 99.8       & 97.8  & 99.1  & 100   & 97.9  & 98.7  & 98.7    & 100   & 100   & 99.1            & 100     & 98.7  & 99.4    & $\uparrow$1.50    & $\downarrow$ 1.30 & $\downarrow$ 0.60  \\ 
\hline
Phoneme                                    & 81.3      & 78.4  & 80.4  & 79.4       & 81.1  & 80.2  & 81.1  & 77.8  & 78.5  & 79.9    & 83.7  & 77.6  & 78.2            & 79.8    & 79.6  & 79.1    & $\downarrow$ 0.30 & $\downarrow$ 1.00 & $\downarrow$ 1.50  \\ 
\hline
Pima                                       & 75.3      & 77.3  & 75.2  & 76.3       & 77.9  & 77.9  & 72.7  & 77.7  & 76.4  & 76.5    & 79.2  & 84.4  & 66.6            & 76.6    & 76.3  & 76.6    & $\uparrow$ 5.90   & $\uparrow$ 1.90   & $\uparrow$ 1.70    \\ 
\hline
Saheart                                    & 77.4      & 73.2  & 70.9  & 72.1       & 75.2  & 67.7  & 79.3  & 69.5  & 64.1  & 71.2    & 72.3  & 73.9  & 65.2            & 69.5    & 71.7  & 71.4    & $\uparrow$ 1.40   & $\downarrow$ 0.40 & $\uparrow$ 0.90    \\ 
\hline
Thyroid                                      & 85.7      & 83.6  & 78.8  & 81.3       & 80.9  & 78.5  & 85.7  & 85.3  & 80.5  & 82.2    & 66.6  & 71.4  & 95.2            & 66.6    & 60.0  & 76.9    & $\uparrow$ 3.50   & $\uparrow$ 0.00   & $\downarrow$ 4.70  \\ 
\hline
Spambase                                   & 93.8      & 92.5  & 93.5  & 93.0       & 93.3  & 93.5  & 93.1  & 92.8  & 93.8  & 93.3    & 93.4  & 93.1  & 95.2            & 93.4    & 94.1  & 93.7    & $\uparrow$ 0.20   & $\uparrow$ 0.20   & $\uparrow$ 0.50    \\ 
\hline
SPECTHeart                                 & 85.1      & 79.1  & 73.6  & 76.3       & 83.3  & 83.3  & 83.1  & 79.2  & 71.6  & 80.1    & 85.1  & 74.1  & 74.1            & 84.6    & 65.3  & 75.6    & $\uparrow$ 9.30   & $\uparrow$ 11.9   & $\uparrow$ 7.50    \\ 
\hline
Tic-Tac-Toe                                & 77.1      & 71.8  & 67.2  & 69.5       & 69.7  & 77.6  & 73.9  & 70.6  & 72.2  & 72.8    & 79.2  & 79.2  & 80.2            & 66.6    & 78.9  & 74.1    & $\uparrow$ 8.00   & $\uparrow$ 8.40   & $\uparrow$ 9.40    \\ 
\hline
Titanic                                    & 75.6      & 77.4  & 78.1  & 77.8       & 73.4  & 80.4  & 79.7  & 78.4  & 78.1  & 78.1    & 78.2  & 80.0  & 76.8            & 77.3    & 76.8  & 77.9    & $\uparrow$ 0.30   & $\uparrow$ 1.00   & $\uparrow$ 1.10    \\ 
\hline
Wdbc                                       & 99.1      & 97.8  & 97.1  & 97.5       & 98.2  & 99.1  & 98.2  & 98.2  & 96.4  & 98.1    & 98.2  & 98.2  & 94.7            & 98.2    & 96.4  & 97.7    & $\uparrow$ 1.90   & $\uparrow$ 1.70   & $\uparrow$ 1.30    \\ 
\hline
Wisconsin                                  & 97.2      & 96.5  & 96.5  & 96.5       & 97.1  & 97.8  & 97.1  & 97.1  & 94.2  & 96.7    & 97.1  & 98.5  & 95.7            & 95.7    & 95.6  & 96.1    & $\uparrow$ 1.10   & $\uparrow$ 1.00   & $\uparrow$ 0.40    \\
\hline
\end{tabular}
\end{adjustbox}
\end{table*}
\begin{figure*}[!htbp]
    \centering
    \begin{subfigure}[b]{0.3\textwidth}
        \centering
        \includegraphics[width=\textwidth]{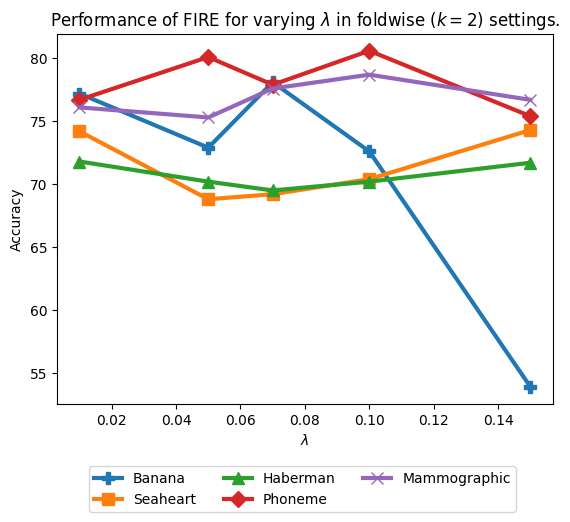}
        \caption{FIRE in folds ($k$ = 2)}
        \label{fig:sub1}
    \end{subfigure}
    \hfill
    \begin{subfigure}[b]{0.3\textwidth}
        \centering
        \includegraphics[width=\textwidth]{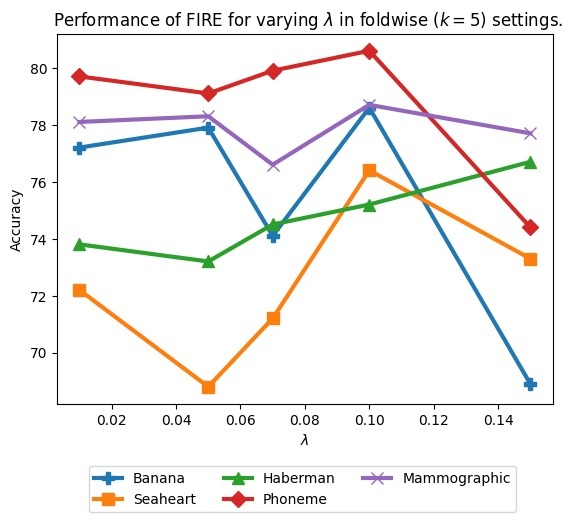}
        \caption{FIRE in folds ($k$ = 5)}
        \label{fig:sub2}
    \end{subfigure}
    \hfill
    \begin{subfigure}[b]{0.3\textwidth}
        \centering
        \includegraphics[width=\textwidth]{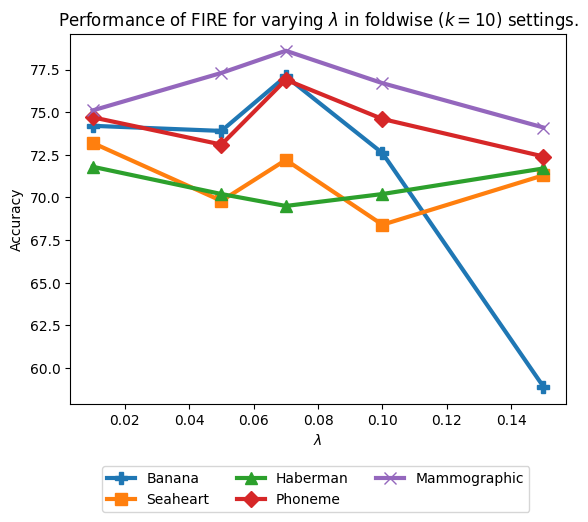}
        \caption{FIRE in folds ($k$ = 10)}
        \label{fig:sub3}
    \end{subfigure}
    \caption{st-CV and FIRE, $\Delta$ accuracy for varying number of folds.}
    \label{fig:main}
\end{figure*}
\begin{figure}
    \centering
    \includegraphics[width=0.5\textwidth]{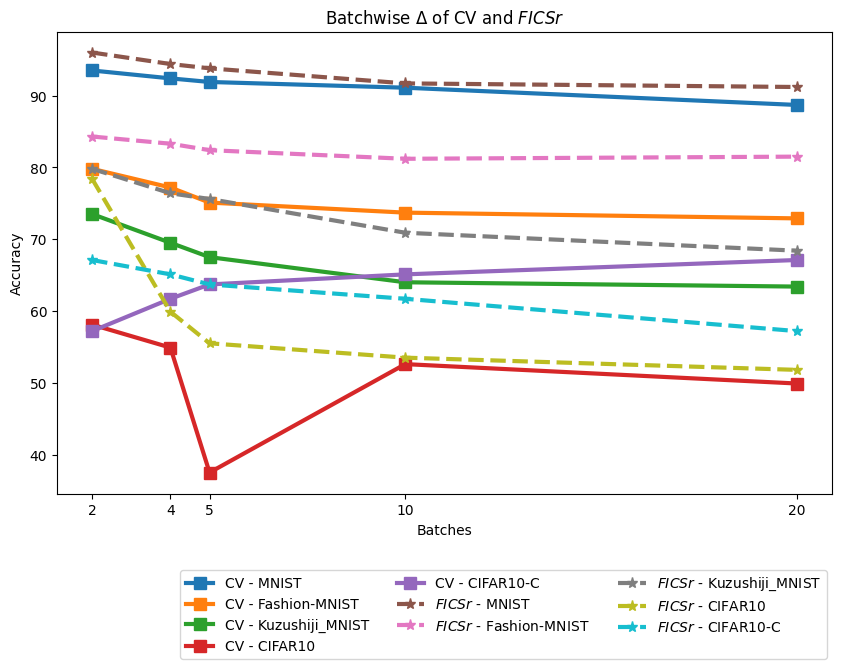}
    \caption{Effect of batching frequency. As the number of batch frequency increases the drop in accuracy also increases.}
    \label{fig:batchfrequency}
\end{figure}

\end{document}